\documentclass[runningheads]{llncs}
\usepackage[T1]{fontenc}
\usepackage{graphicx}
\usepackage{tikz}
\usetikzlibrary{positioning,arrows.meta}
\usepackage{booktabs}
\usepackage{amsmath}
\usepackage{amssymb}
\usepackage{subcaption}
\usepackage{tabularx}
\usepackage[colorlinks=true, allcolors=blue]{hyperref}
\usepackage[nameinlink,noabbrev]{cleveref}
\usepackage{multirow}
\usepackage{xcolor}
\usepackage{eurosym}
\usepackage{tikz}
\usepackage{amssymb}
\usepackage{algorithm}
\usepackage{algpseudocode}
\usepackage{listings}
\usepackage{acro}
\usepackage{xspace}

\acsetup{first-style=long-short}
\DeclareAcronym{llm}{
  short = LLM,
  long = large language model,
  short-plural-form = LLMs,
  long-plural-form = large language models
}
\DeclareAcronym{apo}{
  short = APO,
  long = automatic prompt optimization
}
\DeclareAcronym{LLMaaJ}{
  short = LLMaaJ,
  long = LLM-as-a-Judge
}

\newcommand{\PromptOps}{\textsc{PromptOps}\xspace}
\newcommand{\fairbotbench}{\textsc{FairBotBench}\xspace}

\usepackage{afterpage}

\begin{document}
\raggedbottom

\title{PromptOps: Iterative Prompt Optimization for LLM-as-a-Judge Systems}

\author{
Sven Kleinkop\inst{1} \and
Tamás Janusko\inst{1} \and
Ricardo Bochnia\inst{1} \and
Gunnar Hempel\inst{1} \and
Steffen Tomschke\inst{2} \and
Jürgen Anke\inst{1} \and
Maik Thiele\inst{1}\orcidID{0000-0002-1665-977X}
}
\authorrunning{Kleinkop et al.}
\institute{
  Hochschule für Technik und Wirtschaft Dresden\\
  \email{firstname.lastname@htw-dresden.de}
  \and Otto Group one.O%, Freiberger Str. 35, 01067 Dresden
  \email{firstname.lastname@og1o.de}
  }
\maketitle

\begin{abstract}
\Acp{llm} are increasingly used as automated judges for evaluating the behavior of chatbots and other AI systems. In deployment, however, their agreement with the intended human evaluation policy may degrade as user interactions and score distributions change. Existing automatic prompt optimization methods are primarily studied as one-shot procedures on static development sets and therefore provide limited guidance for maintaining \ac{llm} judges over time.
We frame this problem as \emph{judge drift} and introduce \PromptOps, a closed-loop framework for monitoring and iteratively recalibrating \ac{LLMaaJ} systems. \PromptOps separates four operational decisions: when to optimize, which optimization procedure to apply, which feedback records to use, and how to update the judge prompt. We instantiate the framework with five prompt optimization methods: GEPA, SIMBA, Rubric Synthesis, and exemplar selection using either simulated annealing or a Tree-structured Parzen Estimator.
Our experiments demonstrate that iterative prompt optimization can recalibrate \ac{llm} judges and recover performance after judge drift. However, effective maintenance is inherently context-dependent: neither optimization algorithm choice nor update timing generalizes universally across settings. 
These findings establish prompt optimization for deployed LLM judges as an inherently iterative maintenance problem and demonstrate that \PromptOps provides a practical abstraction for studying and implementing this process.

\end{abstract}

\section{Introduction}
Large language models (\acp{llm}) are increasingly deployed in chatbots and agentic assistants.
As reliance on these systems grows, scalable and trustworthy evaluation of their outputs becomes increasingly important.
\ac{LLMaaJ} systems address this need by using one or more \acp{llm} to score, rank, or critique model outputs according to natural-language criteria~\cite{li2025generation,gu2026survey}.
This flexibility is particularly useful for behavioral and safety evaluation, where judges must recognize subtle forms of toxicity, bias, inappropriate personalization, and other undesirable behavior.
However, strong agreement with human evaluations on a fixed benchmark does not imply that this agreement will persist after deployment.

A judge that is well aligned with human evaluations at one point in time may lose this agreement as the stream of evaluated interactions changes.
We refer to this phenomenon as \emph{judge drift}: a loss of agreement between a fixed judge and the intended human evaluation policy under a changing evaluation stream.
Judge drift may arise from changes in user behavior, application policies, or the distribution of encountered cases.
Importantly, this notion of drift does not require the judge model itself to change; drift can emerge solely from changes in the stream of cases being evaluated. Maintaining judge alignment under such changes turns prompt design from a one-time configuration task into a continuing maintenance problem.

Human-centered and practitioner workflows already describe evaluation design as an iterative process in which people define criteria, inspect disagreements, and extend test sets~\cite{pan2024human,shankar2024validators,lee2026dataprompt}.
At the same time, prompt revisions are not guaranteed to improve behavior globally: changes that correct known failures can introduce regressions on previously well-handled cases~\cite{commey2026betterprompts}.
Most work on \ac{apo}, however, studies optimization as a one-shot procedure on a static development set.
A deployed \ac{LLMaaJ} system instead requires repeated decisions about \emph{when} recalibration is necessary, \emph{which} observations should inform it, and \emph{how} the judge prompt should be updated under practical cost constraints.

To address this problem, we introduce \PromptOps, a closed-loop framework for the iterative monitoring and recalibration of \ac{LLMaaJ} systems.
\PromptOps separates the operational decisions involved in judge maintenance into four components:
a \emph{Controller} determines when optimization should be triggered, a \emph{Dispatcher} selects the optimization procedure to execute, a \emph{Sample Selector} determines which accumulated feedback records are used for optimization, and an \emph{Optimizer} proposes an updated judge prompt.
This modularization allows prompt-optimization algorithms to be studied not only in isolation, but as components of a recurring feedback loop in which alignment improvements must be balanced against the cost of optimization. This shifts prompt optimization from a one-shot algorithmic problem to an operational control problem in which adaptation quality, timing, and cost must be jointly managed.

Specifically, we make the following contributions:
\begin{itemize}
    \item We formulate \emph{judge drift} as a deployment-time maintenance problem for \ac{LLMaaJ} systems and provide a controlled experimental setting for studying adaptation under changing evaluation streams.
    \item We introduce \PromptOps, a modular closed-loop framework that makes optimization triggering, optimizer selection, feedback-record selection, and prompt updating independently configurable and evaluable.
    \item We instantiate \PromptOps with five prompt-optimization procedures and show across two judge backends that iterative recalibration can recover performance after induced drift, while both optimizer choice and recalibration timing exhibit substantial backend- and cost-dependent trade-offs.
\end{itemize}

\section{Related Work}
\subsection{LLM-as-a-Judge}
The \ac{LLMaaJ} paradigm uses instruction-tuned \acp{llm} as flexible evaluators across a wide range of criteria~\cite{li2025generation,gu2026survey}.
Prompted general-purpose models have emerged as an alternative to learned evaluation metrics such as BARTScore and UniEval \cite{yuan2021bartscore,zhong2022unieval,wang2023chatgpt,fu2024gptscore}.
% TODO: Den nächsten Satz überarbeiten, der ist noch nicht rund
Benchmarks such as MT-Bench and Chatbot Arena established their use for model comparison~\cite{zheng2023mtbench}, while G-Eval demonstrated improved agreement with human ratings through explicitly generated evaluation steps~\cite{liu2023geval}.
Despite these advances, human agreement, reliability, robustness, and systematic judge biases remain central challenges~\cite{gu2026survey}.

Recent work therefore focuses on improving the evaluation procedure itself.
CHECK-EVAL and RocketEval decompose evaluation criteria into explicit checklists~\cite{pereira2024checkeval,wei2025rocketeval}, RevisEval constructs response-adapted references~\cite{zhang2025reviseval}, and inversion learning has been used to derive task-specific evaluation prompts from single examples~\cite{hong2025inversion}.
These approaches improve individual evaluation procedures, whereas we study how prompt-improvement methods can be repeatedly orchestrated within an operational feedback loop.

\subsection{Human Alignment of LLM Judges}
A central quality criterion for an \ac{llm} judge is its agreement with the intended evaluation policy, which is commonly operationalized through human reference annotations.
Throughout this paper, \emph{human alignment} refers narrowly to agreement with task-specific human judgments, rather than to the broader behavioral or safety alignment of the underlying model.
Because human judgments can themselves vary, judge agreement should be interpreted relative to inter-annotator agreement whenever such a baseline is available.

Prior work has documented several systematic deviations from human judgments.
These include position bias, where the ordering of candidate responses affects pairwise decisions~\cite{shi2025positionbias,zheng2023mtbench}; verbosity or fluency bias, where stylistic properties influence ratings independently of task quality~\cite{zhou2024evaluationbias,zheng2023mtbench}; and self-preference, where evaluators favor outputs generated by the same model family~\cite{panickssery2024selfpreference}.
Such vulnerabilities are not eliminated simply by selecting a more capable judge model~\cite{thakur2025judges}.

Human-centered approaches therefore retain people in the process of specifying evaluation criteria, reviewing failures, and updating evaluation policies~\cite{pan2024human,shankar2024validators}.
This perspective also treats datasets and prompts as coupled artifacts: newly validated cases inform prompt revisions, while revised prompts can in turn expose further failure cases~\cite{lee2026dataprompt}.
Existing work predominantly studies such disagreement and its mitigation either in static settings, or through the lens of discrete human intervention.
We instead focus on maintaining judge alignment, both automatically and continuously, as feedback accumulates and the distribution of evaluated interactions changes over time.

\subsection{Automatic Prompt Optimization}
Prompt-improvement methods differ in the level of access they require to the target model.
White-box techniques may exploit model parameters, gradients, or token probabilities, whereas discrete black-box techniques treat the model as a queryable service and optimize prompts based on generated outputs or externally computed scores~\cite{ramnath2025aposurvey}.
We focus on the latter setting because it reflects many API-based judge deployments.

Black-box \ac{apo} searches over instructions, rubrics, demonstrations, or other prompt components to improve an externally measured objective.
Even small changes in wording, ordering, and formatting can substantially affect model behavior~\cite{sclar2024sensitivity}.
Evolutionary search constitutes one prominent family of black-box \ac{apo} methods.
EvoPrompt instantiates genetic and differential-evolution strategies using \ac{llm}-generated mutations~\cite{guo2025evoprompt}, while GEPA combines reflective prompt evolution with Pareto-aware candidate selection~\cite{agrawal2026gepa}.
Related work has examined mutation operators and early-stopping strategies for improving the efficiency of evolutionary prompt search~\cite{griesshaber2025toolbox}.
HPSS specializes genetic search to \ac{llm} judges by decomposing a judge prompt into dimensions such as evaluation criteria, scoring scale, and component order~\cite{wen2025hpss}.

A complementary line of work seeks to reduce annotation or evaluation cost.
Active-sampling \ac{apo} selects uncertain and diverse records for human labeling~\cite{zhen2025activesampling}, while uncertainty-aware comparative optimization can reduce the number of pairwise evaluations required on human-annotated targets~\cite{fathullah2025uncertainty}.
Prompt Duel Optimizer instead formulates label-free prompt comparison as a dueling-bandit problem, although its optimization signal remains dependent on the quality of the evaluating \ac{llm}~\cite{wu2026pdo}.

Most \ac{apo} evaluations optimize a prompt once against a fixed development set.
Our contribution is orthogonal to any particular optimization algorithm:
\PromptOps embeds prompt optimization in a recurring feedback loop over the accumulated history $\mathcal{H}_t$ and separates decisions about when optimization is triggered, which records are selected, and which Optimizer or sequence of Optimizers is executed.

\section{PromptOps Framework}
\label{sec:promptops_framwork}
\begin{figure}
    \centering
    \includegraphics[width=0.95\linewidth]{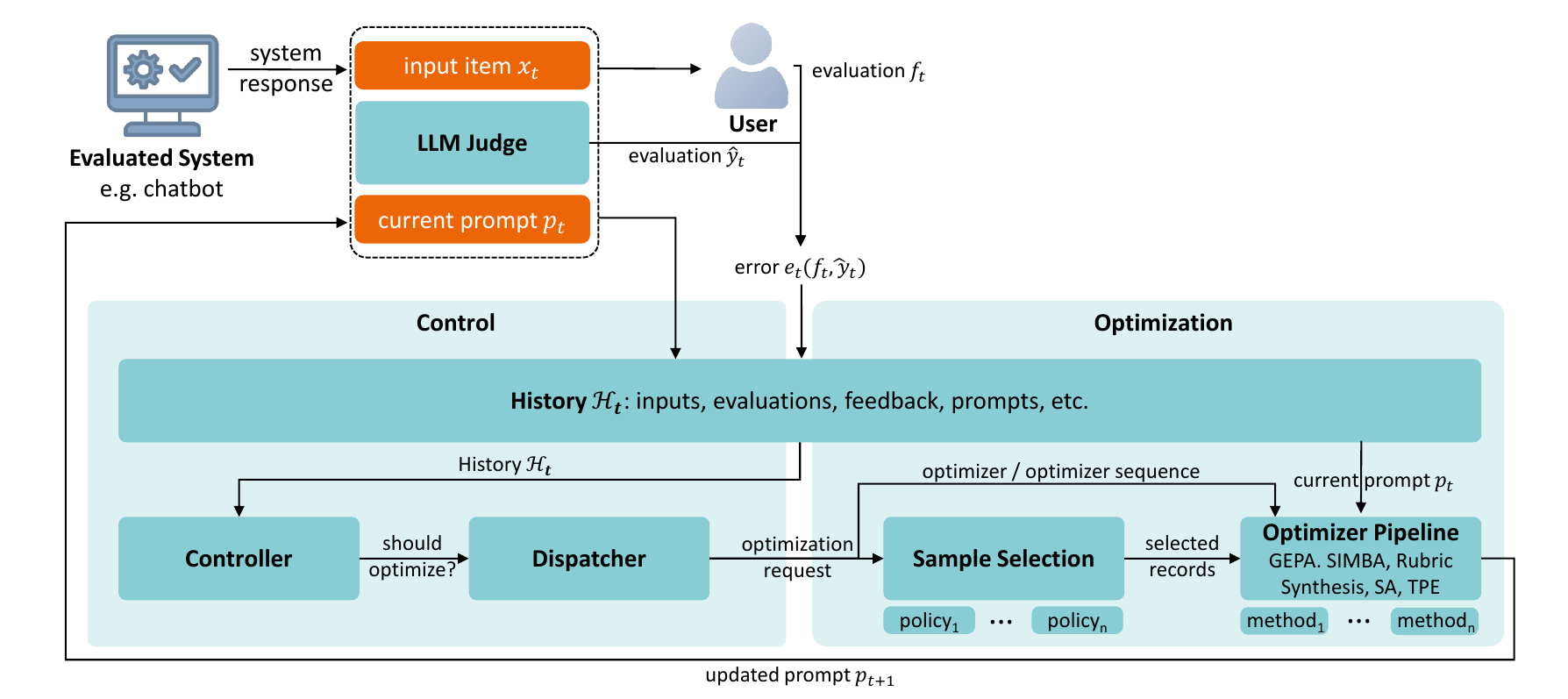}
    \caption{Overview of \PromptOps as a closed-loop extension of an LLM judge. At time $t$, the judge evaluates input item $x_t$ using prompt $p_t$ and produces evaluation $\hat{y}_t$. When reference feedback $f_t$ is available, the resulting error $e_t$ and all evaluation artifacts are accumulated in history $H_t$. The Controller monitors this history, while the Dispatcher, Sample Selector, and Optimizer determine whether and how a new prompt $p_{t+1}$ is produced.}
    \label{fig:framework-overview}
\end{figure}
\PromptOps extends an \ac{LLMaaJ} system with a continuous feedback loop for monitoring and recalibrating its evaluation behavior. Rather than treating the judge prompt as a static configuration, the framework maintains an evolving prompt $p_t$ that can be updated as new evaluations and reference feedback accumulate. At time $t$, the judge applies $p_t$ to an input item $x_t$ and produces an evaluation, including a numerical score $\hat{y}_t$. When reference feedback $f_t$ becomes available, it is added to the accumulated history $\mathcal{H}_t$, which provides the basis for subsequent recalibration decisions. Figure~\ref{fig:framework-overview} summarizes this closed-loop process and its main components. Note that feedback $f_t$ does not necessarily belong to the same domain as $\hat{y}_t$, e.g. binary user evaluation ´´thumbs up´´ / ´´thumbs down´´, or written feedback. 
%When a reference score $y_t$ becomes available, the resulting feedback is added to the accumulated history $\mathcal{H}_t$, which provides the basis for subsequent recalibration decisions. Figure~\ref{fig:framework-overview} summarizes this closed-loop process and its main components.

The accumulated history $\mathcal{H}_t$ holds all relevant information, including past input items, evaluation artifacts, all feedback, dispatch \& optimization records, as well as past prompts.
The Controller monitors this history and decides whether to initiate an optimization pass, in which case the Dispatcher selects one or more optimization procedures, the corresponding Sample Selector chooses which records to use for prompt generation and validation, and the Optimizer proposes $p_{t+1}$.
The operational objective is therefore not only to minimize predictive loss, but to do so under a budget for model calls, tokens, latency, and human feedback.

In our experiments, dataset reference annotations serve as an idealized proxy for deployment feedback.
This choice isolates the behavior of the optimization loop, but does not reproduce the sparsity, delay, noise, disagreement, or acquisition cost of real user feedback.
Our experiments therefore study adaptation under oracle-like feedback rather than realistic label acquisition.
\subsection{Optimizers}
The \emph{Optimizer} is responsible for updating the current judge prompt. \PromptOps does not prescribe a particular optimization method, but defines a common interface through which an Optimizer receives the current prompt and selected records from the accumulated feedback history and returns a candidate updated prompt. We evaluate five base optimization algorithms: GEPA, SIMBA, Rubric Synthesis, and exemplar selection using either simulated annealing (SA) or a Tree-structured Parzen Estimator (TPE):\\\\
\textbf{GEPA} is an evolutionary \ac{apo} method that uses language-model reflection on execution traces to propose revisions of the current prompt $p_t$~\cite{agrawal2026gepa}.
Within \PromptOps, these traces are derived from the records selected from $\mathcal{H}_t$ by the Sample Selector.
GEPA maintains a population of candidate prompts and uses Pareto-aware selection to preserve candidates that perform strongly on different subsets of the selected records rather than reducing all observations to a single aggregate score.
Candidates are initially evaluated on minibatches, while promising candidates receive broader evaluation on the available validation records.
%The resulting candidate selected by GEPA becomes the updated prompt $p_{t+1}$.
GEPA is available in prompt-optimization frameworks including DSPy~\cite{khattab2023dspy}\footnote{\url{https://dspy.ai/learn/optimization/optimizers/}} and DeepEval\footnote{\url{https://deepeval.com/docs/prompt-optimization-introduction}}.\\
\textbf{SIMBA} iteratively derives candidate rules and demonstrations from execution trajectories associated with previously observed records.
Within \PromptOps, SIMBA operates on the records provided by the Sample Selector and uses them to generate and evaluate candidate revisions of $p_t$.
Its candidate population is updated through repeated minibatch evaluation. During each step SIMBA either adds a new rule to the candidate or updates its list of exemplars. %with the selected candidate yielding the updated prompt $p_{t+1}$.
SIMBA was introduced as part of the DSPy framework\footnote{\url{https://dspy.ai/api/optimizers/SIMBA/}}.
Singhvi et al.\ report that SIMBA slightly outperformed GEPA on medical-imaging benchmarks~\cite{singhvi2025prompttriage}, motivating its inclusion as a complementary optimization strategy despite the substantial difference in evaluation domain.\\
\textbf{Rubric Synthesis} is a lightweight custom optimization method that restricts revisions of the current prompt $p_t$ to its scoring rubric, while leaving all other prompt components unchanged.
This design is consistent with prior approaches that focus on generating detailed or adaptive evaluation criteria~\cite{liu2024autocalibrate,fan2025sedareval}.
During each optimization pass, the Sample Selector provides a bounded set of feedback records from the accumulated history $\mathcal{H}_t$.
Conditioned on these records and the current rubric, an expert \ac{llm} generates twelve candidate rubrics: six are synthesized from scratch, while six revise the current rubric.
Each candidate rubric is combined with the unchanged remainder of $p_t$, yielding twelve candidate prompts.
Together with the unmodified prompt $p_t$, these thirteen candidates are evaluated on a common development minibatch.
The highest-scoring candidate is retained as the updated prompt $p_{t+1}$.
Bounding the number of records exposed to the expert \ac{llm} keeps its optimization context focused and the associated inference cost predictable.\\\\
A complementary optimization strategy is exemplar selection, which modifies the current prompt \(p_t\) through demonstrations rather than instruction or rubric revision. Given a pool of candidate exemplars obtained from the selected records in $\mathcal{H}_t$, we search the combinatorial space of possible exemplar subsets and insert the selected subset into the judge prompt to produce $p_{t+1}$. We consider two methods for searching over candidate exemplar subsets: simulated annealing and a Tree-structured Parzen Estimator: Simulated annealing (SA) explores this search space by iteratively adding, removing, or replacing a single exemplar.
To escape local optima, it may probabilistically accept a lower-scoring candidate, with the acceptance probability decreasing over the course of the search~\cite{kirkpatrick1983annealing}.
The Tree-structured Parzen Estimator (TPE) instead treats exemplar selection as a hyperparameter optimization problem.
It models promising and less promising configurations separately and uses these models to estimate which exemplar subsets are most likely to improve the optimization objective~\cite{bergstra2011tpe}.

We include exemplar-selection Optimizers because prior work suggests that exemplar optimization remains comparatively underexplored relative to instruction optimization and can provide complementary improvements~\cite{wan2024exemplars}.
Our TPE-based variant follows a similar principle to the exemplar-selection procedures employed by SIMBA and MIPROv2 \cite{opsahlong2024}.
\subsection{Dispatcher}
When the Controller triggers an optimization pass, the Dispatcher determines which Optimizer or sequence of Optimizers to execute.
A single-method Dispatcher always invokes the same Optimizer, whereas a composite Dispatcher applies multiple Optimizers sequentially, with each method operating on the prompt produced by its predecessor.
More generally, dispatch policies may also rotate among a predefined set of Optimizers or adapt their selection based on the cost and outcomes of previous optimization passes.

In this work, we evaluate single-method dispatch and fixed composite configurations. For the latter, Optimizer order is fixed.
%part of the dispatch policy, as changing the execution order can result in different updated prompts.
%
\subsection{Controller}
The Controller monitors the accumulated history $\mathcal{H}_t$ and determines when to initiate an optimization pass.
Because $\mathcal{H}_t$ contains the observed inputs, evaluation artifacts, feedback, and previous optimization records, a wide range of triggering policies can be defined.
In this work, we consider two such policies: 
The \emph{periodic policy} triggers optimization at increasing intervals. Given a period $p$, optimization is first triggered at $t_0 = p$. Subsequent passes follow at steps $t_{n+1}$ defined by Equation~\ref{eq:period}.
\begin{equation}
    t_{n+1}
    =
    \left\lceil
        t_n + p \cdot 1.5^{n+1}
    \right\rceil,
    \label{eq:period}
\end{equation}
%
%$t_k=\lceil t_0 r^k\rceil$, where $t_0$ determines the initial interval and $r>1$ progressively increases the time between successive optimization passes.
The \emph{performance-based policy} computes an error statistic $E_t$ over a sliding window containing the most recent $w$ feedback records from $F=\{ f_{\hat{t}} \in \mathcal{H}_t \mid \hat{t} \le t \}$ and triggers an optimization pass whenever $E_t\ge\tau$ for a predefined threshold $\tau$.
After a trigger, the window is reset.
Consequently, at least $w$ new feedback records must be accumulated before another optimization can be initiated, making $w$ also serve as a cooldown period to prevent overly frequent optimization passes.
\section{Experiments}
We structure our evaluation around three questions.
First, we examine the error--cost trade-offs of different Dispatcher configurations under stationary data streams (\Cref{subsec:general_evaluation}).
Second, we investigate whether iterative prompt optimization can recover judge performance after an induced shift in the reference-score distribution (\Cref{subsec:dataset-adaptability}).
Third, we study how a performance-based Controller affects the timing and effectiveness of recalibration compared with periodic triggering (\Cref{subsec:sliding-window-Controller}).
\subsection{Experimental Setup}
\paragraph{Data and metric.}
We evaluate \PromptOps on \fairbotbench{}, which comprises 600 German--English e-commerce conversations with human reference annotations across three dimensions: bias, toxicity, and personalization~\cite{fairbotbench}.
Each run processes a stream of 500 conversation turns, and each experimental condition is repeated on three separately sampled streams to reduce sensitivity to individual stream samples.
We follow the iterative procedure outlined by \PromptOps.
%We follow a streaming evaluation protocol: each turn is evaluated using the judge prompt available at that point in the stream, and its reference feedback is added to the optimization history only after the evaluation.
Consequently, during prompt optimization, only observations encountered so far may be accessed, never future ones.

For each turn, we average the three associated human reference annotations on the $1$--$5$ scale and use the resulting value as reference feedback. Our primary error measure is mean absolute error (MAE), where lower values indicate closer agreement between judge predictions and human reference scores.

The initial judge prompt is generated using the DeepEval implementation of G-Eval%
\footnote{\url{https://deepeval.com/docs/metrics-llm-evals}}.
Following the procedure described by \cite{liu2023geval}, we provide a description of the evaluation criteria from which the evaluation steps are synthesized.
The resulting prompt serves as the common starting point for all \PromptOps runs.
When applied without subsequent optimization, it also defines the static baseline used in the drift experiments.
The input criteria and the generated G-Eval prompt are documented in Appendix~\ref{app:initial-prompt}.
\paragraph{Models.}
We use two judge backends and one expert model:
GPT-5.4-nano with low reasoning effort serves as the proprietary judge backend, while GPT-5.4 with high reasoning effort serves as the expert model used during prompt optimization.
As an open-weight judge backend, we use the \path{gemma-4-26B-A4B-it-UD-Q4_K_M} quantization of Gemma 4%
\footnote{\url{https://huggingface.co/unsloth/gemma-4-26B-A4B-it-GGUF}}.
The OpenAI models are accessed through the OpenAI API, whereas Gemma 4 is served locally using the \texttt{llama.cpp} server module with default settings.

For each condition, we record cumulative token usage to characterize the computational cost of the different Dispatcher configurations.
Because tokenization differs across model families, token counts are compared only between configurations using the same judge backend.
\paragraph{Dispatcher, Sample Selector, and Optimizer.}
We consider a total of 13 Dispatcher pipelines in this work comprising the five Optimizers described in Section~\ref{sec:promptops_framwork}. Before the main evaluation, we tune Optimizer parameters on a small stream of 49 turns from \fairbotbench{}, disjoint from those used during evaluation. Our default Sample Selector policy, \emph{NewElements}, uses observations received since the preceding optimization pass as training records and the training records from the preceding pass as the development set. Detailed descriptions of Dispatcher configurations, Optimizer parameters, and Sample Selectors are reported in Appendix~\ref{app:optimizer-parameters}.
\paragraph{Controller.}
Unless stated otherwise, experiments use a deterministic periodic Controller with $p=20$.
This schedule places optimization passes more densely near the beginning of the stream and increasingly sparsely thereafter.
For a stream of $500$ items, optimization passes are thus scheduled, following Equation~\ref{eq:period}, at steps $t \in \{20, 50, 95, 163, 265, 417\}$.
Together with the initial G-Eval prompt, these six passes yield at most seven distinct prompt versions per run.
An optimization pass does not necessarily change the prompt, as an optimizer may retain the current prompt if it does not identify an improvement.
\subsection{Optimizer Screening and Cost--Error Trade-offs}
\label{subsec:general_evaluation}
We first screen 23 parameterized configurations spanning 13 Dispatcher pipelines.
% Referenz auf Tabelle im Anhang eher nicht, oder? listed in Table~\ref{tab:optimizer-configurations}.
The additional configurations result from small, medium, and large parameterizations of Rubric Synthesis and SIMBA, and two parameterizations each of GEPA and TPE.
An exhaustive cross-product of Optimizer parameterizations within composite pipelines was infeasible under the available time and token budgets.
We therefore use predominantly small variants in composite configurations, with \textit{TPE(L)} and \textit{Rubric(M)} included as comparatively inexpensive larger variants.

\begin{figure}[!htbp]
  \centering
  \begin{subfigure}{\textwidth}
    \centering
    \makebox[\linewidth][c]{\includegraphics[width=1.10\linewidth]{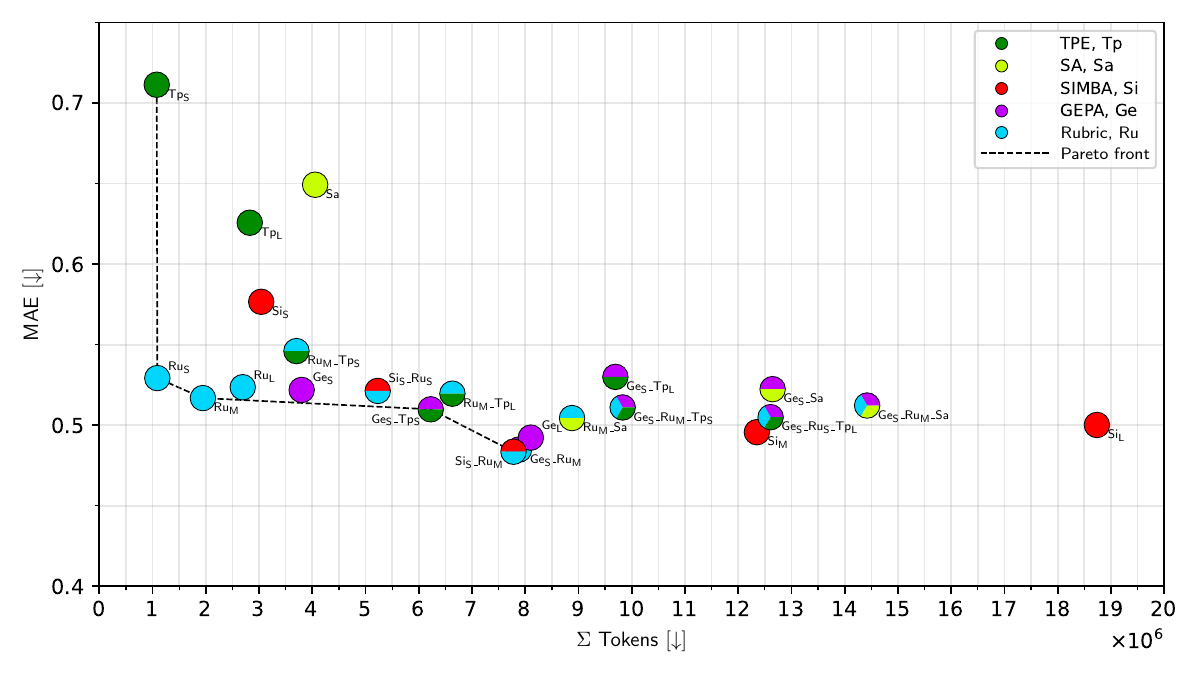}}
    \caption{GPT judge}
    \label{fig:pareto-screening-gpt}
  \end{subfigure}
  %\vspace{0.1em}
  \begin{subfigure}{\textwidth}
    \centering
    \makebox[\linewidth][c]{\includegraphics[width=1.10\linewidth]{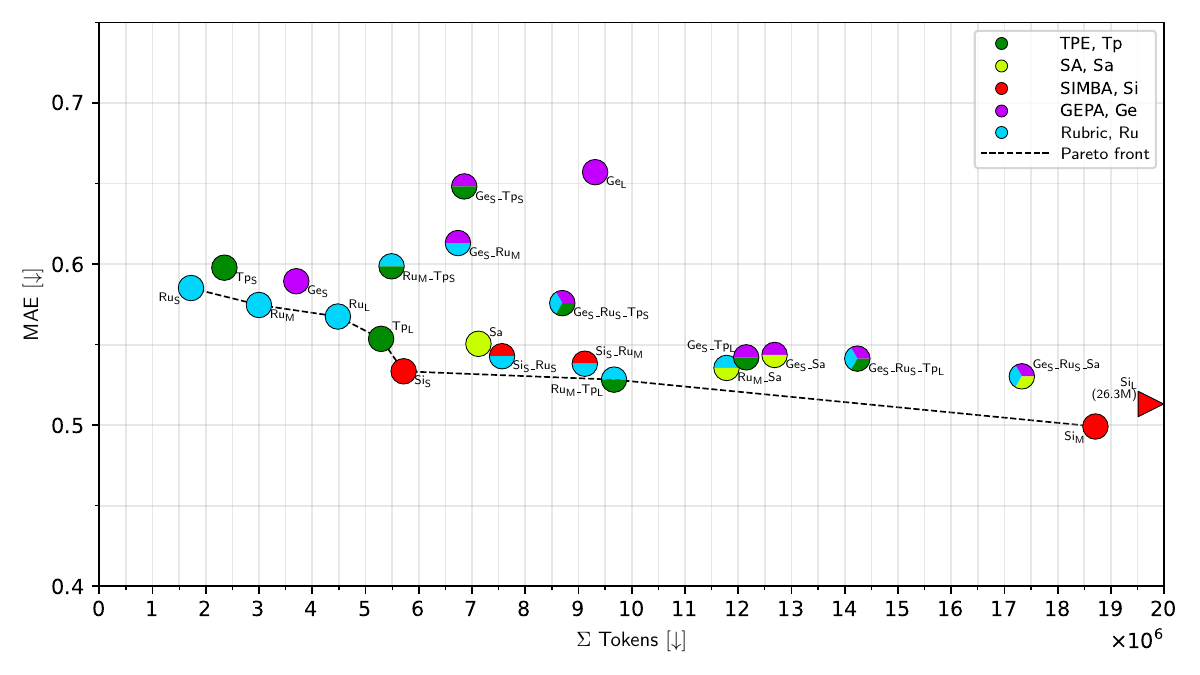}}
    \caption{Gemma 4 judge}
    \label{fig:pareto-screening-gemma}
  \end{subfigure}
  \caption{Mean MAE and cumulative token usage for 23 parameterized Dispatcher configurations across three repeated runs, shown separately for the two judge backends.
  Each marker represents one configuration; colors indicate the Optimizers used, with multicolored markers denoting composite pipelines.
  Dashed lines connect the non-dominated configurations on the Pareto front.
  Lower MAE and lower token usage correspond to more favorable operating points.}
  \label{fig:pareto-screening}
\end{figure}

Figure~\ref{fig:pareto-screening} summarizes the trade-off between mean MAE and cumulative token usage across three repeated runs for each configuration.
All 23 configurations are evaluated separately with the GPT and Gemma 4 judge backends.
Because prompt optimization is stochastic and each condition is repeated only three times, we use this experiment as a screening study rather than to establish a definitive ranking of Optimizer configurations.
Corresponding plots based on median MAE are reported in Appendix~\ref{app:median-pareto}.

For the GPT judge, the Pareto front ranges from inexpensive TPE(L) and Rubric configurations to the GEPA(S)$\rightarrow$\allowbreak TPE(S) composite and the SIMBA(S)$\rightarrow$\allowbreak Rubric(M) composite.
The latter achieves the lowest observed mean MAE of approximately 0.48, while several configurations consume more tokens without reducing the error further.
For Gemma 4, the shape of the Pareto front differs substantially.
All three Rubric parameterizations occupy the low-token region of the front, which then extends to SIMBA(M), with approximately 18.7 million tokens and a mean MAE of 0.5.
SIMBA(L) uses approximately 26.3 million tokens without improving on SIMBA(M) and is therefore dominated.
%Out of the 6 configurations with the highest MAE, 4 include the GEPA Optimizer, with both pure GEPA configurations performing noticably worse, relatively speaking, compared to the OpenAI backend.

Overall, the backend-specific Pareto fronts indicate that the relative cost-effectiveness of Dispatcher configurations does not transfer directly between judge models.
%In particular, configurations that provide favorable MAE--token trade-offs for one backend are not necessarily Pareto-optimal for the other.
In particular, only two configurations are Pareto-optimal across both backends: Rubric(S) and Rubric(M). 
%One major difference between two judge models is the performance of configurations employing only exemplar-selections. TPE(S), TPE(L), and SA performed the worst out of all configurations for GPT models, while yielding substantially better results with Gemma judges.    
%Based on the backend-specific mean and median Pareto analyses, we retain six configurations for the subsequent drift experiments for each judge backend. For the GPT backend: Rubric(S), Rubric(L), TPE(S), GEPA(S), GEPA(S)$\rightarrow$\allowbreak TPE(S), and SIMBA(S)$\rightarrow$\allowbreak Rubric(M). For the Gemma backend: Rubric(S), Rubric(L), TPE(L), SIMBA(S), SIMBA(M), SIMBA(S)$\rightarrow$\allowbreak Rubric(M). The latter was chosen over Rubric(M)$\rightarrow$\allowbreak TPE(L), because of its median performance and to match the GPT selection. 
%
\subsection{Adaptation under Induced Judge Drift}
\label{subsec:dataset-adaptability}
\begin{figure}[t]
  \centering
  \begin{subfigure}[b]{0.325\linewidth}
    \centering
    \includegraphics[width=\linewidth]{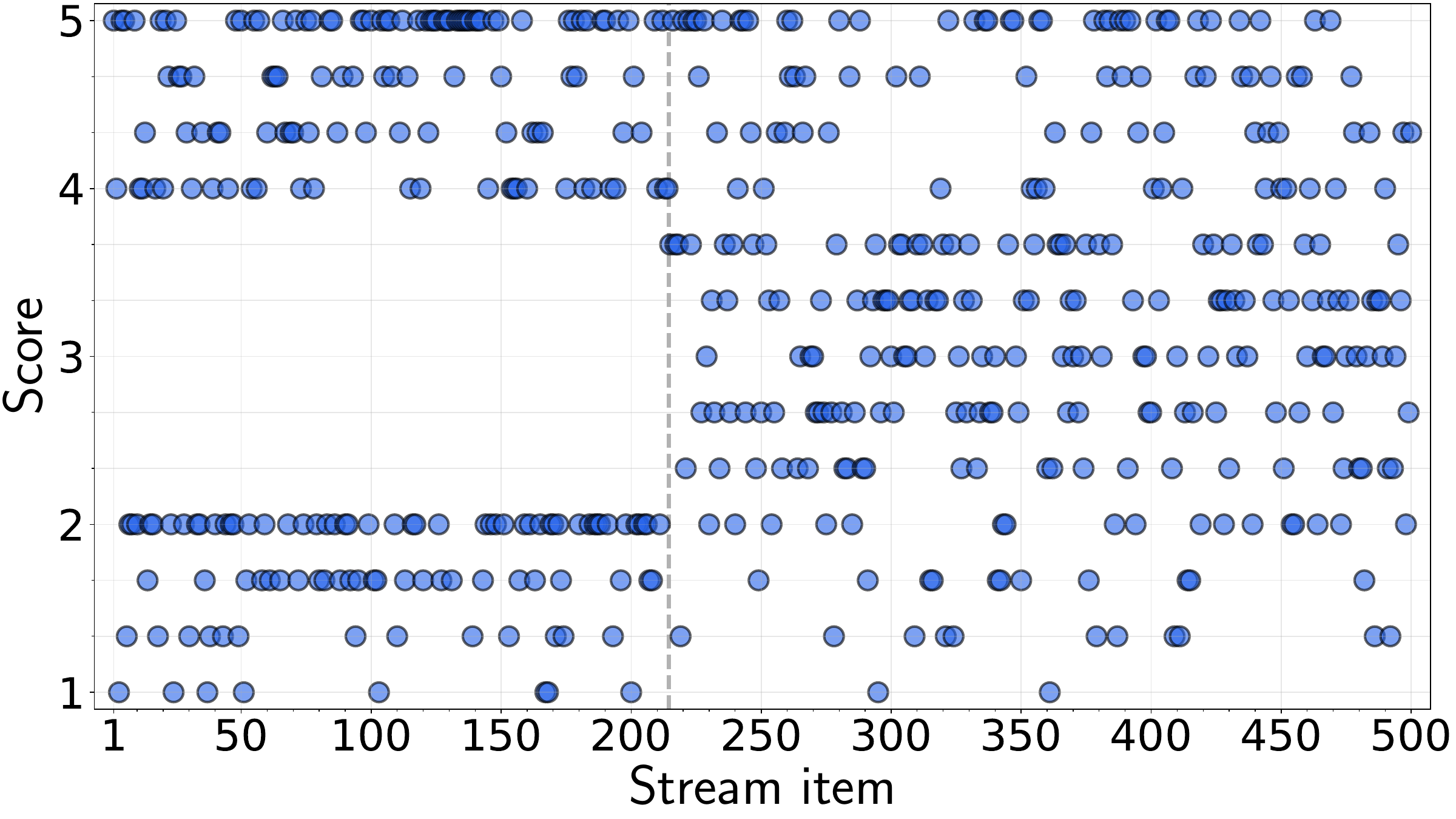}
    \caption{Easy.}
    \label{fig:drift-streams-easy}
  \end{subfigure}\hfill
  \begin{subfigure}[b]{0.325\linewidth}
    \centering
    \includegraphics[width=\linewidth]{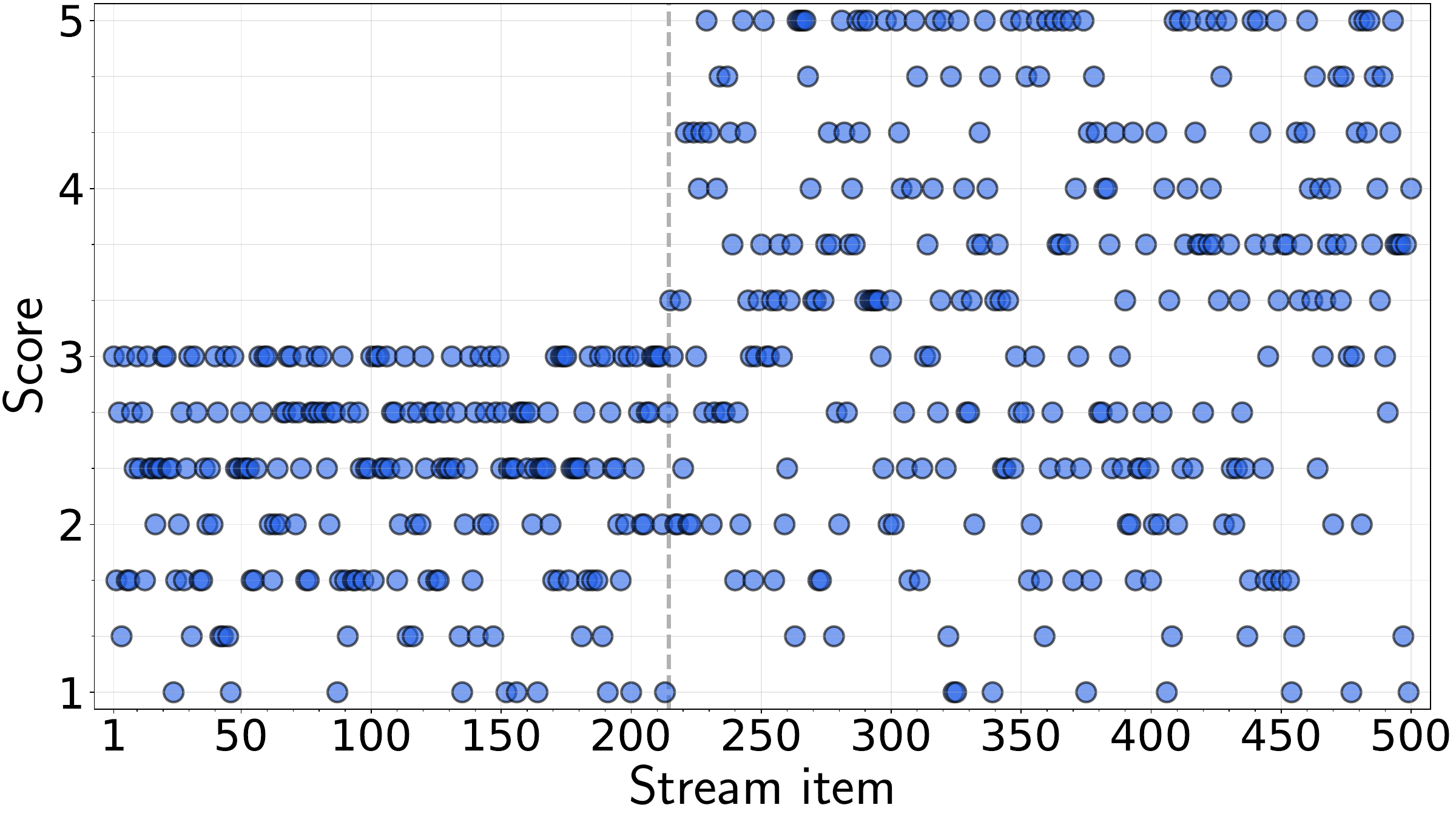}
    \caption{Low--First.}
    \label{fig:drift-streams-low-only}
  \end{subfigure}\hfill
  \begin{subfigure}[b]{0.325\linewidth}
    \centering
    \includegraphics[width=\linewidth]{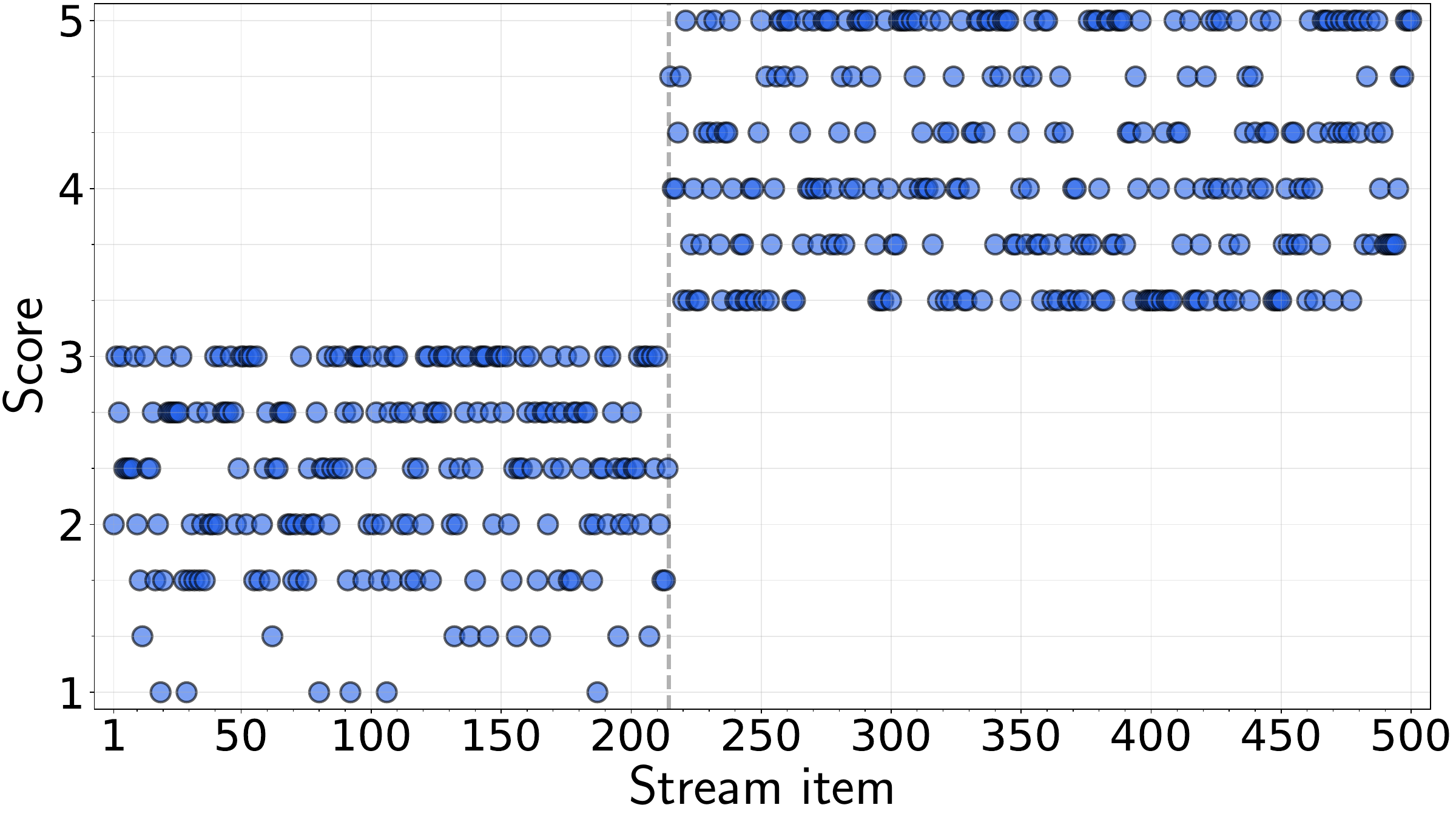}
    \caption{Low--High Split.}
    \label{fig:drift-streams-low-high}
  \end{subfigure}
  \caption{Reference-score streams used to evaluate adaptation under an induced distribution shift.
  Each point represents the averaged human reference score at the corresponding stream position.
  The dashed vertical line marks the shift boundary after item 214.}
  \label{fig:drift-streams}
\end{figure}

The uniformly sampled streams used in the screening preserve the reference-score distribution of \fairbotbench{} and therefore do not introduce an explicit distribution shift.
We construct three different types of streams with an abrupt change after item 214, as shown in Figure~\ref{fig:drift-streams}.
The shift boundary lies between the periodic optimization passes at items 163 and 265, leaving a post-shift interval before the next scheduled recalibration.
%

\begin{figure}[!htbp]
  \centering
  \begin{subfigure}{0.49\linewidth}
    \centering
    \includegraphics[width=\linewidth]{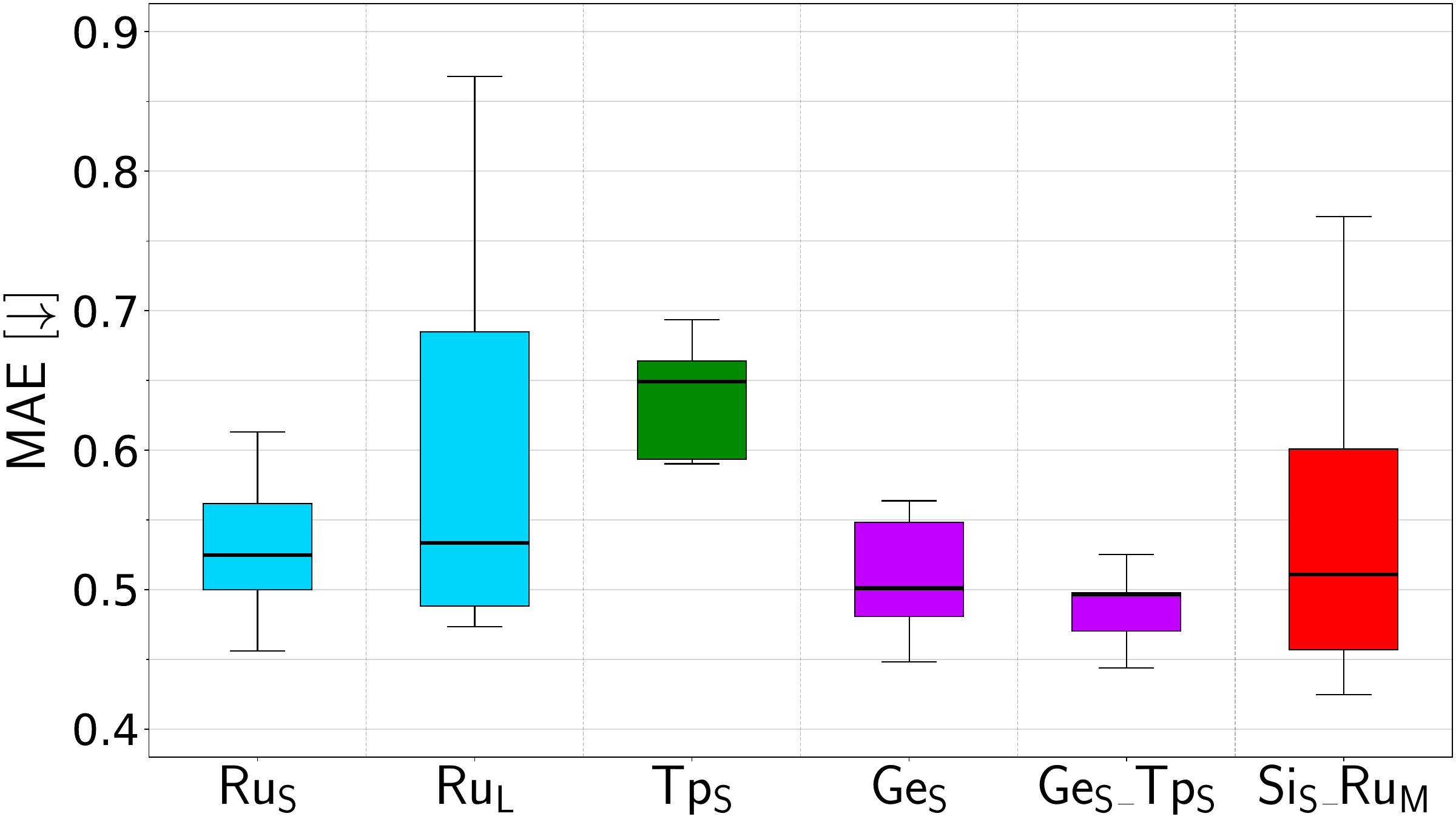}
    \caption{GPT}
    \label{fig:drift-overall-gpt}
  \end{subfigure}\hfill
  \begin{subfigure}{0.49\linewidth}
    \centering
    \includegraphics[width=\linewidth]{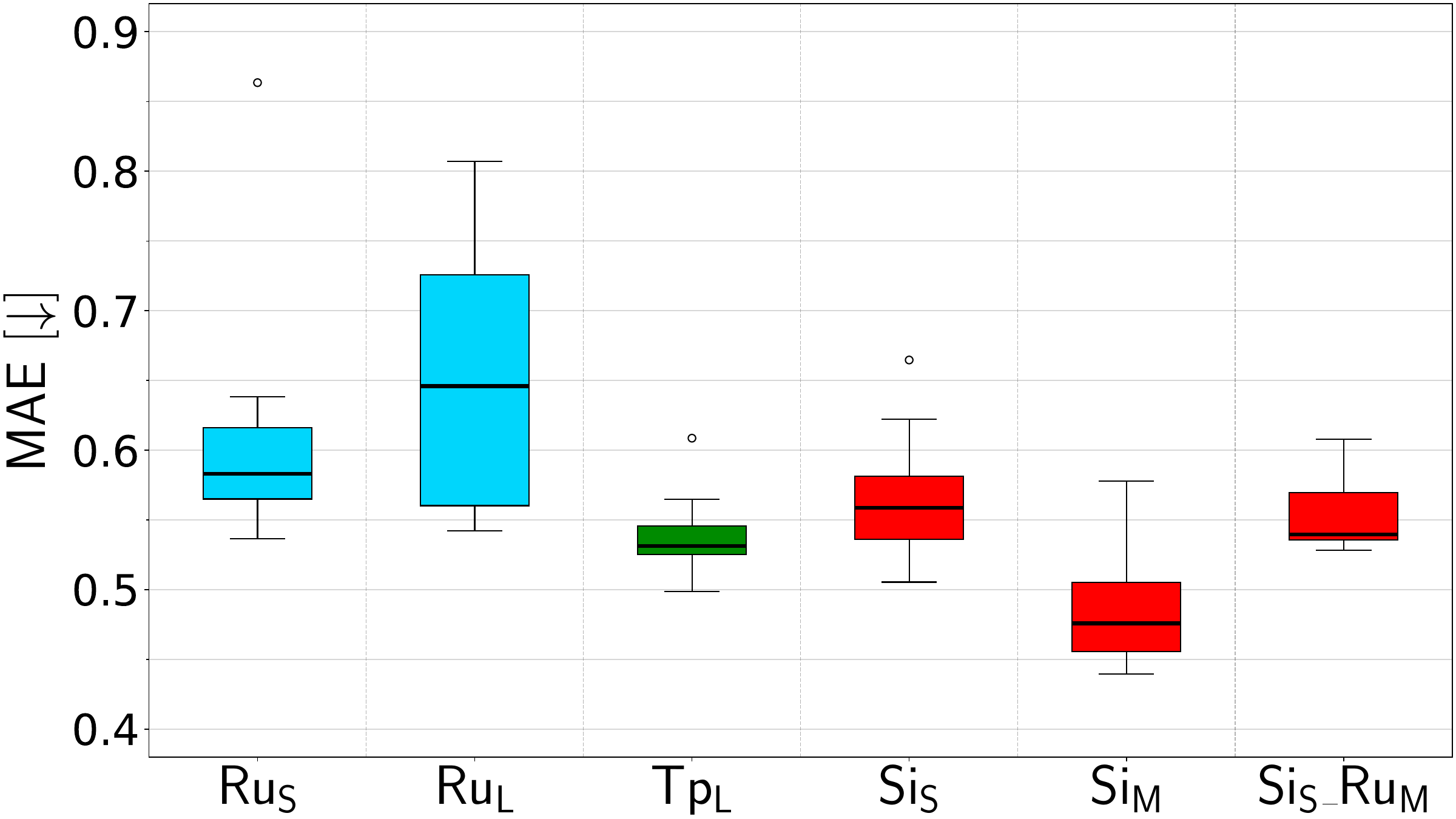}
    \caption{Gemma 4}
    \label{fig:drift-overall-gemma}
  \end{subfigure}
  \caption{Overall MAE (lower is better) of the six retained Dispatcher configurations across nine drift runs per model backend (three stream types and three samples per type). Center lines denote medians; boxes and whiskers summarize between-run variation.}
  \label{fig:drift-overall}
\end{figure}

\begin{figure}[!htbp]
  \centering
  \begin{subfigure}{0.49\textwidth}
    \centering
    \includegraphics[width=\linewidth]{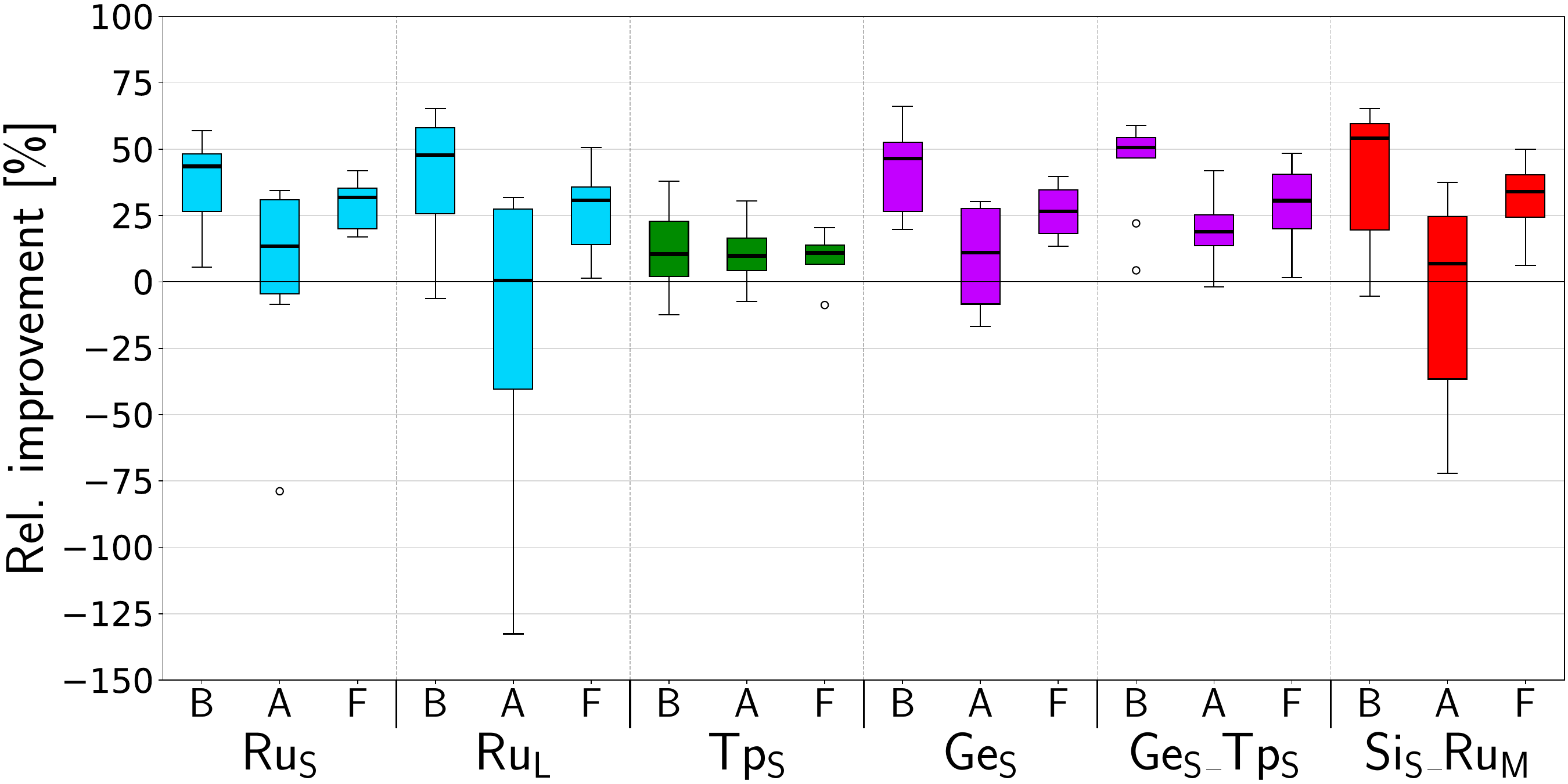}
    \caption{GPT}
    \label{fig:drift-intervals-gpt}
  \end{subfigure}\hfill
  \begin{subfigure}{0.49\textwidth}
    \centering
    \includegraphics[width=\linewidth]{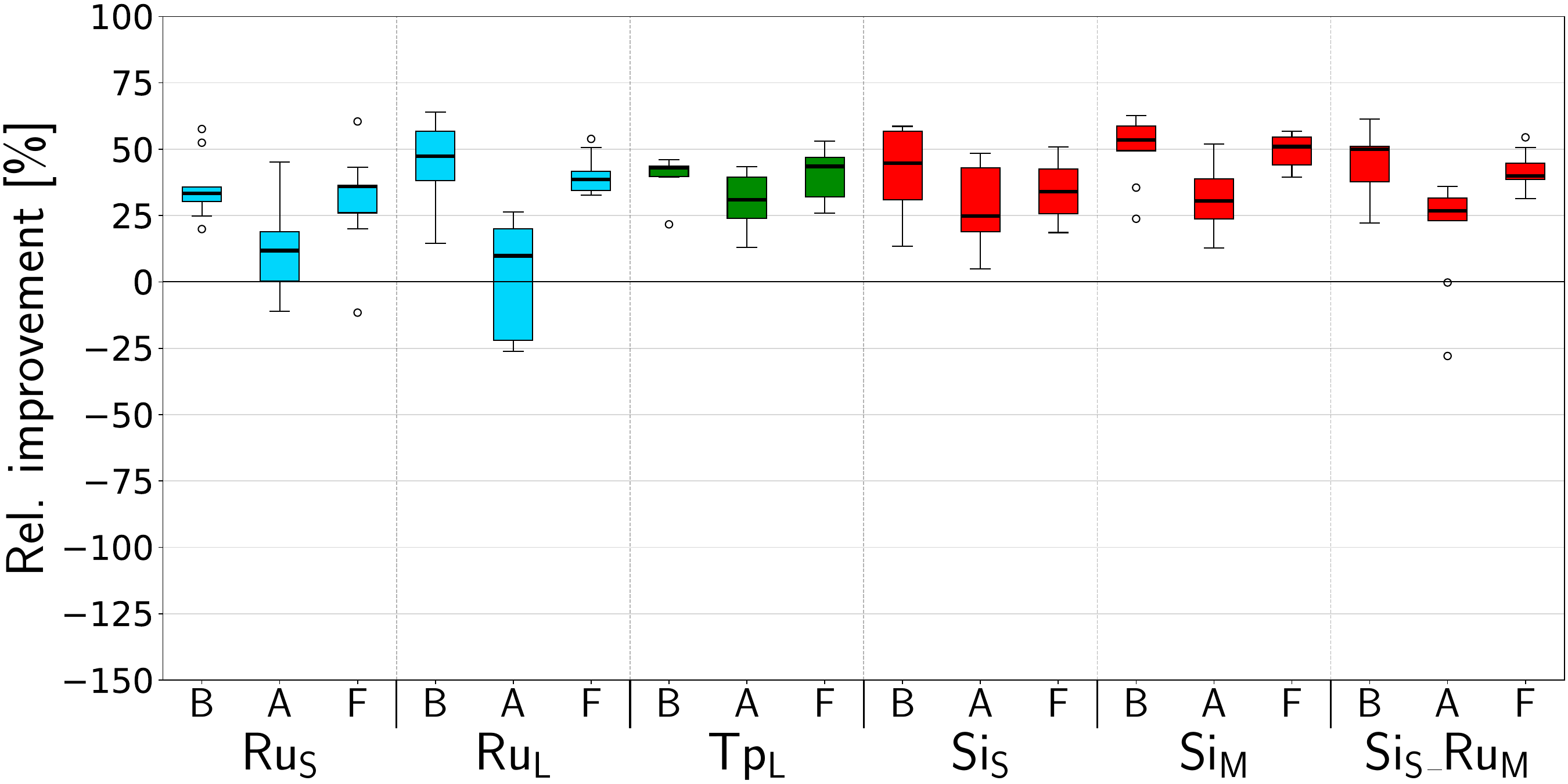}
    \caption{Gemma 4}
    \label{fig:drift-intervals-gemma}
  \end{subfigure}
  \caption{Improvement over the static initial-prompt baseline before the distribution shift ($[164,214]$), immediately after the shift ($[215,265]$), and after the final optimization pass ($[418,500]$).
  Boxes pool nine runs across three stream types, with three sampled streams per type.
  Positive values indicate lower MAE than the static baseline; center lines denote medians.
  }
  \label{fig:drift-intervals}
\end{figure}

%We evaluate each of the six retained Dispatcher configurations on three sampled streams of each type,
Based on the backend-specific mean and median Pareto analyses\footnote{Note that SIMBA(S)$\rightarrow$\allowbreak Rubric(M) was chosen over Rubric(M)$\rightarrow$\allowbreak TPE(L) because of its median performance and to match the GPT selection. 
}, we retain six configurations per backend, and evaluate each on three sampled streams of each type, yielding nine runs per configuration and judge backend.
Figure~\ref{fig:drift-overall} provides a descriptive overview of the resulting overall MAE.
For the GPT judge, GEPA(S)$\rightarrow$\allowbreak TPE(S) exhibits the lowest median MAE and the narrowest interquartile range, whereas TPE(S) has the highest median MAE. Rubric(L) and SIMBA(S)$\rightarrow$\allowbreak Rubric(M) show the highest variation across the nine runs.
Rubric(L) also retains its high variation, when using the Gemma backend, and yields the weakest results in that setting.
In contrast, for the Gemma judge, Simba(M) performs best, achieving substantially lower median MAE compared to the other five evaluated configurations, while TPE(L) exhibits the narrowest interquartile range.
Because the three stream types differ in their reference-score distributions, their absolute MAE values are not directly comparable.
We therefore complement the overall results with improvement relative to the static baseline, which applies the initial G-Eval prompt without subsequent optimization.
For an experimental run, relative improvement is computed as
$$
  100 \cdot
  \frac{
    \operatorname{MAE}_{\mathrm{base}}
    -
    \operatorname{MAE}_{\mathrm{exp}}
  }{
    \operatorname{MAE}_{\mathrm{base}}
  }.$$
Positive values indicate lower MAE than the static baseline.
Figure~\ref{fig:drift-intervals} reports this measure before the distribution shift, immediately after the shift, and after the final optimization pass.

Immediately after the distribution shift, relative improvement decreases for all twelve Dispatcher configurations, and individual runs of the Rubric and SIMBA(S)$\rightarrow$\allowbreak Rubric(M) configurations fall below the static baseline with both backends.
By the final interval, however, all twelve configurations again show consistent improvement over the baseline.
For most configurations, the final median remains below its corresponding pre-shift value. Rubric(S) and TPE(L) of the Gemma backend are the exception and reach their highest median improvement after the final iteration.
We therefore interpret the observed behavior as recovery after the distribution shift rather than complete restoration of pre-shift performance. 
%The relative scale is important here: the stream constructions differ in difficulty, and on the pre-shift segment of Easy the static prompt remains competitive with optimized prompts.
%This difference in difficulty, however, might also apply to the post-shift segments when compared to their corresponding pre-shift segments, due to the changes to their score distribution. 
%

Table~\ref{tab:low-filtered-results} provides a more detailed view of the \emph{Low--First} streams, including the interval following the first scheduled post-shift optimization pass, $[266,417]$.

For the GPT judge, Rubric(L) and SIMBA(S)$\rightarrow$\allowbreak Rubric(M) perform strongest before the shift, both reaching a rounded median MAE of 0.37.
After the distribution shift, however, GEPA(S)$\rightarrow$\allowbreak TPE(S) is substantially more robust, achieving the lowest immediate post-shift MAE of 0.49 and remaining the strongest configuration after the first scheduled recalibration.
By the final interval, SIMBA(S)$\rightarrow$\allowbreak Rubric(M) shows the strongest recovery, reaching an MAE of 0.46 and a 34.0\% improvement over the static baseline.

For the Gemma judge, Rubric(L) and SIMBA(M) exhibit the lowest MAE before the shift with 0.34 an 0.35, respectively. After the shift, SIMBA(S) achieves the highest relative improvement with 43.0\% over the baseline. The lowest MAE, however, is reached by TPE(L), with 0.53. TPE(L) in general remains remarkably consistent with all of its MAE values falling in the range $[0.53, 0.57]$. 

SIMBA(M) performs best on the two optimization intervals, as well as overall with an MAE of 0.51. However, SIMBA(S), SIMBA(S)$\rightarrow$\allowbreak Rubric(M), and TPE(L) remain close with MAEs not exceeding 0.54. 
\begin{table}[!htbp]
  \caption{Median MAE (abs.) and median improvement over the static initial-prompt baseline (rel., \%) on the \emph{Low--First} streams. The intervals are overall $[1,500]$, pre-shift $[164,214]$, immediate post-shift $[215,265]$, first post-shift optimization $[266,417]$, and final $[418,500]$. Positive relative values indicate lower MAE than the baseline. Arrows indicate an increase or decrease in relative improvement from the preceding reported interval; bold values mark the best rounded result in each column. The upper block reports GPT results, while the lower block contains results using Gemma 4 judges.
  }
  \label{tab:low-filtered-results}
  \centering
  \resizebox{\textwidth}{!}{\begingroup
\renewcommand{\arraystretch}{1.2} % Vertical padding
\setlength{\tabcolsep}{8pt}       % Horizontal padding per column side
\newcommand{\trendup}{\textcolor{green!60!black}{\ensuremath{\uparrow}}}
\newcommand{\trenddown}{\textcolor{red}{\ensuremath{\downarrow}}}
\newcommand{\trendvalue}[2]{%
  \makebox[3em][r]{#1}%
  \makebox[1em][c]{#2}%
}
\begin{tabular}{lcc|cccc|cccc}
\toprule
 & \multicolumn{2}{c|}{Overall} & \multicolumn{2}{c}{Pre-shift} & \multicolumn{2}{c|}{Post-shift} & \multicolumn{2}{c}{After opt.} & \multicolumn{2}{c}{Final} \\
 & \multicolumn{1}{c}{abs} & \multicolumn{1}{c|}{rel} & \multicolumn{1}{c}{abs} & \multicolumn{1}{c}{rel} & \multicolumn{1}{c}{abs} & \multicolumn{1}{c|}{rel} & \multicolumn{1}{c}{abs} & \multicolumn{1}{c}{rel} & \multicolumn{1}{c}{abs} & \multicolumn{1}{c}{rel} \\
\midrule
\multicolumn{11}{l}{\textit{GPT}} \\
\addlinespace[2pt]
Rubric(S) & 0.54 & 27.3 & 0.38 & 54.3 & 0.67 & \trendvalue{-8.4}{\trenddown} & 0.72 & \trendvalue{-3.9}{\trendup} & 0.48 & \trendvalue{31.8}{\trendup} \\
Rubric(M) & 0.53 & 28.3 & \textbf{0.37} & \textbf{58.0} & 0.66 & \trendvalue{-12.1}{\trenddown} & 0.69 & \trendvalue{10.8}{\trendup} & 0.52 & \trendvalue{24.9}{\trendup} \\
TPE(S) & 0.68 & 8.2 & 0.73 & 10.5 & 0.59 & \trendvalue{5.1}{\trenddown} & 0.58 & \trendvalue{9.3}{\trendup} & 0.65 & \trendvalue{10.9}{\trendup} \\
GEPA(S) & 0.55 & 26.3 & 0.39 & 52.8 & 0.61 & \trendvalue{1.1}{\trenddown} & 0.58 & \trendvalue{16.5}{\trendup} & 0.61 & \trendvalue{16.3}{\trenddown} \\
GEPA(S)$\rightarrow$TPE(S) & \textbf{0.50} & \textbf{33.2} & 0.41 & 50.6 & \textbf{0.49} & \trendvalue{\textbf{20.5}}{\trenddown} & \textbf{0.51} & \trendvalue{\textbf{25.9}}{\trendup} & 0.51 & \trendvalue{27.0}{\trendup} \\
SIMBA(S)$\rightarrow$Rubric(M) & \textbf{0.50} & 28.5 & \textbf{0.37} & 57.9 & 0.61 & \trendvalue{6.8}{\trenddown} & 0.58 & \trendvalue{5.4}{\trenddown} & \textbf{0.46} & \trendvalue{\textbf{34.0}}{\trendup} \\
\midrule
\multicolumn{11}{l}{\textit{Gemma}} \\
\addlinespace[2pt]
Rubric(S) & 0.62 & 32.8 & 0.44 & 52.4 & 0.83 & \trendvalue{0.7}{\trenddown} & 0.57 & \trendvalue{33.5}{\trendup} & 0.62 & \trendvalue{36.1}{\trendup} \\
Rubric(L) & 0.65 & 29.6 & \textbf{0.34} & \textbf{59.2} & 1.04 & \trendvalue{-11.0}{\trenddown} & 0.84 & \trendvalue{16.0}{\trendup} & 0.59 & \trendvalue{38.6}{\trendup} \\
TPE(L)& 0.53 & 39.3 & 0.56 & 39.7 & \textbf{0.53} & \trendvalue{39.5}{\trenddown} & 0.57 & \trendvalue{33.8}{\trenddown} & 0.54 & \trendvalue{32.1}{\trenddown} \\
SIMBA(S) & 0.54 & 41.1 & 0.40 & 56.8 & 0.54 & \trendvalue{\textbf{43.0}}{\trenddown} & 0.55 & \trendvalue{36.6}{\trenddown} & 0.51 & \trendvalue{42.5}{\trendup} \\
SIMBA(M)  & \textbf{0.51} & \textbf{43.7} & 0.35 & 58.7 & 0.58 & \trendvalue{31.3}{\trenddown} & \textbf{0.52} & \trendvalue{\textbf{45.8}}{\trendup} & \textbf{0.47} & \trendvalue{\textbf{51.0}}{\trendup} \\
SIMBA(S)$\rightarrow$Rubric(M) & 0.54 & 36.3 & 0.45 & 51.2 & 0.72 & \trendvalue{23.0}{\trenddown} & 0.55 & \trendvalue{37.2}{\trendup} & 0.58 & \trendvalue{39.4}{\trendup} \\
\bottomrule
\end{tabular}
\endgroup

}
\end{table}
\subsection{Sliding-Window Controller}
\label{subsec:sliding-window-Controller}
The \emph{Low--First} results show that the distribution shift can reverse an improvement and that subsequent periodic passes may be required for recovery.
This effect is most pronounced for the Rubric variants.
We therefore conduct an exploratory evaluation of the performance-based sliding-window Controller defined in Section~\ref{sec:promptops_framwork}.
Due to monetary and local-compute budget constraints, this evaluation uses only GEPA(S) and Rubric(M) Dispatcher configurations on the \emph{Low--First} stream with the GPT backend.

The sliding-window Controller is parameterized by its window $w$ and its threshold $\tau$. We explore two values for each: $w \in \{20, 50\}$ (\emph{Small} and \emph{Big} window) and $\tau \in \{0.4, 0.6\}$ (\emph{High} sensitivity, \emph{Low} sensitivity).
%The Controller computes the MAE over a window of $w$ recent observations and triggers optimization when that MAE exceeds $\tau$.
%We use a small window of $w=20$ or a big window of $w=50$.
%We call $\tau=0.4$, which is 10\% of the four-point score range, \emph{high sensitivity}, and $\tau=0.6$, or 15\% of the range, \emph{low sensitivity}.
Operationally, this Controller policy models a setting in which repeated negative feedback triggers a review of the judge prompt.

\begin{figure}[!htbp]
  \centering
  \begin{subfigure}{0.49\linewidth}
    \centering
    \includegraphics[width=\linewidth]{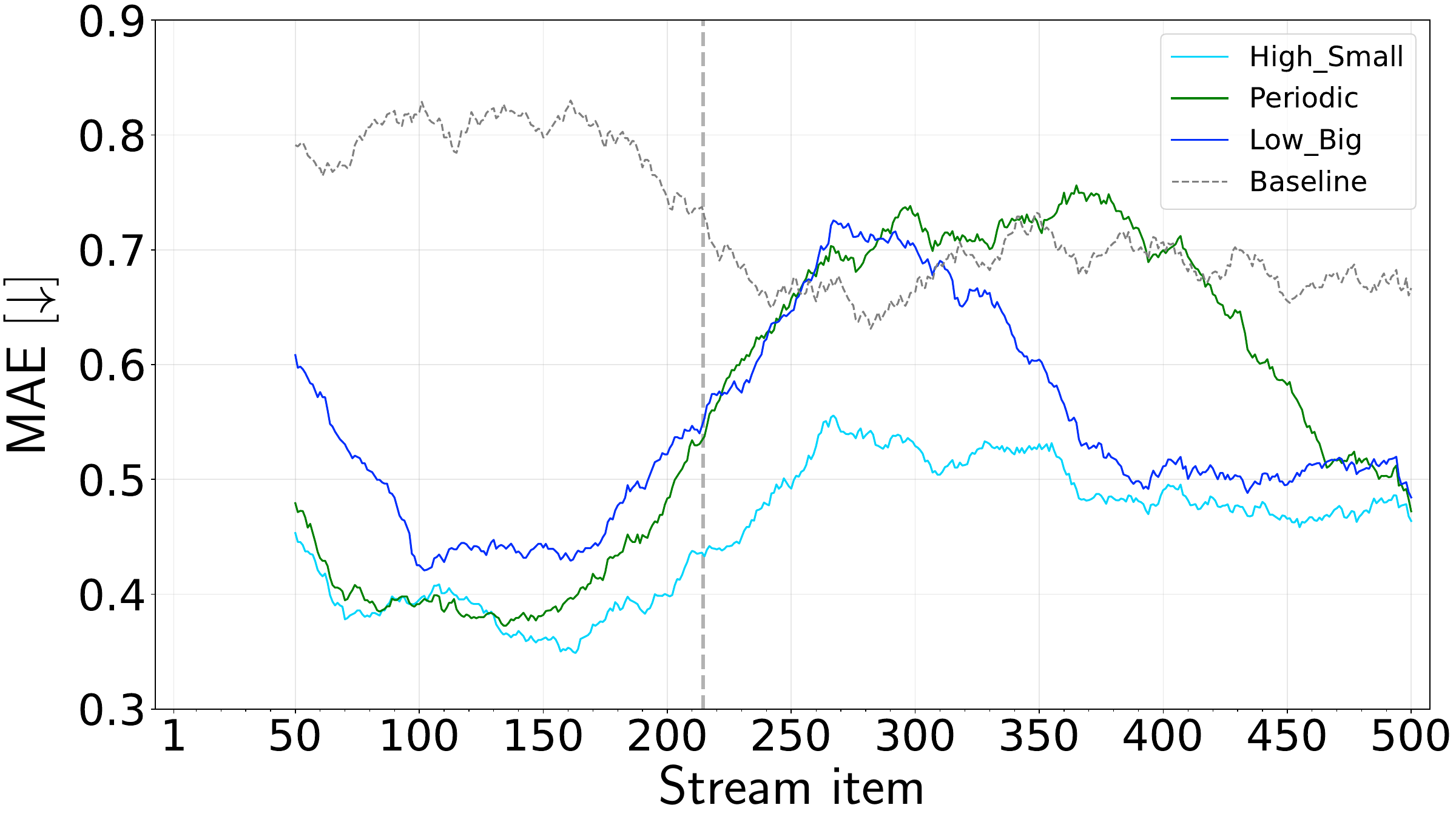}
    \caption{Rubric Synthesis.}
    \label{fig:controller-rolling-rubric}
  \end{subfigure}\hfill
  \begin{subfigure}{0.49\linewidth}
    \centering
    \includegraphics[width=\linewidth]{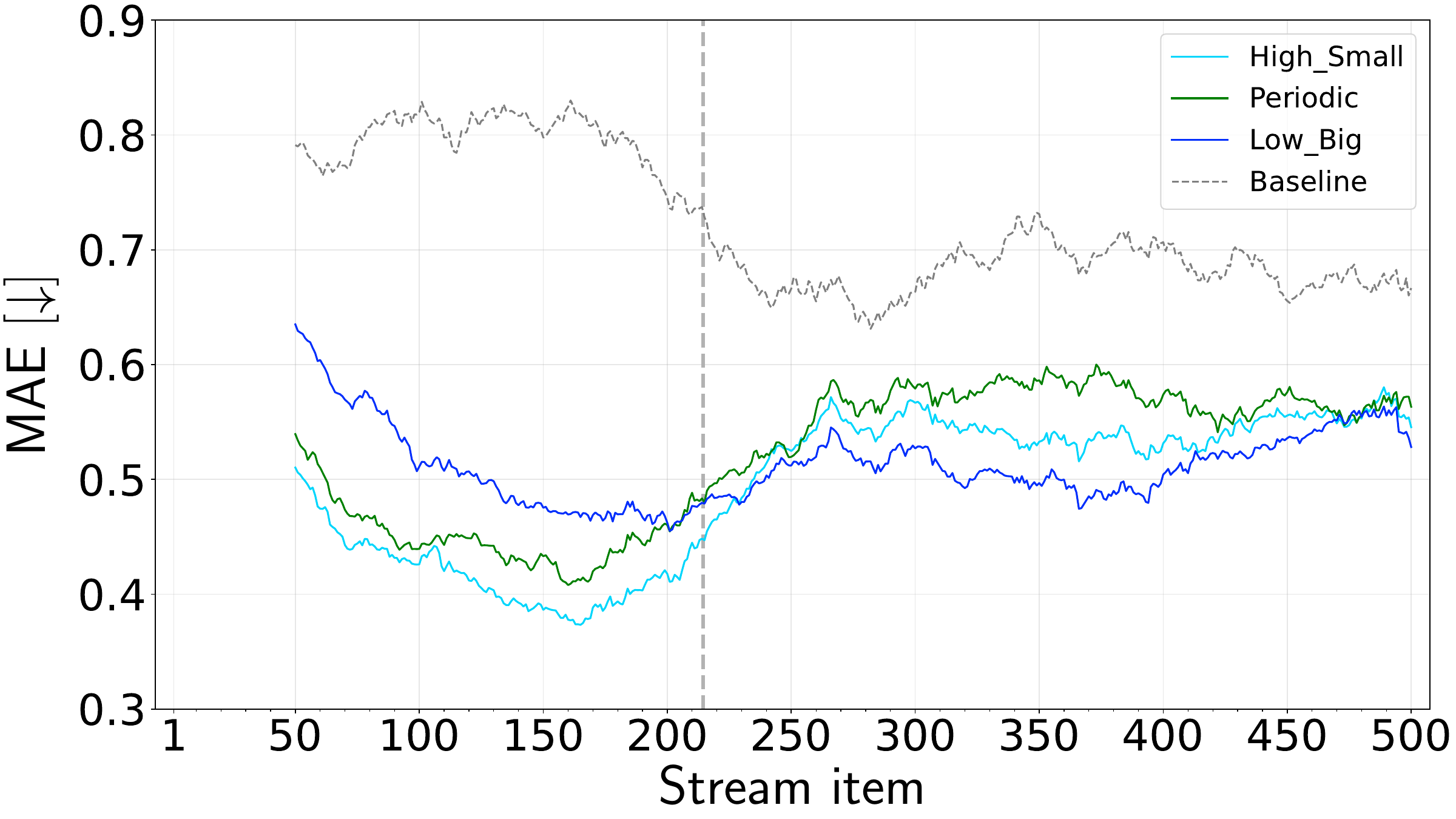}
    \caption{GEPA.}
    \label{fig:controller-rolling-gepa}
  \end{subfigure}
  \caption{Median rolling MAE (lower values are better) computed over a sliding evaluation window of 50 observations, for High--Small, Low--Big, the periodic Controller, and the static initial-prompt baseline. The dotted vertical line marks the distribution boundary, item 214.}
  \label{fig:controller-rolling}
\end{figure}

Table~\ref{tab:controller-results} reports the eight combinations of Dispatcher configuration, window size, and sensitivity.
High--Small achieves the lowest overall MAE for both GEPA (0.49) and Rubric Synthesis (0.47), whereas Low--Big performs worst for both (0.54 and 0.56).
All four GEPA settings improve on its periodic result of 0.59.
For Rubric Synthesis, High--Small and High--Big improve on the periodic result of 0.53, Low--Small ties it at the reported precision, and Low--Big is 0.03 MAE (5.7\%) worse.
Thus, seven of eight adaptive settings match or improve on the corresponding periodic result.
\begin{table}[!t]
  \caption{Median overall MAE on the \emph{Low--First} streams for the performance-based Controller. Big and Small denote sliding windows of 50 and 20 observations; High and Low denote high sensitivity ($\tau=0.4$) and low sensitivity ($\tau=0.6$), respectively. The Periodic row reports the corresponding results from the default scheduled Controller. Lower values are better, and bold values mark the best setting for each Optimizer.}
\medskip
  \label{tab:controller-results}
  \centering
  \begingroup
\renewcommand{\arraystretch}{1.2} % Vertical padding
\setlength{\tabcolsep}{8pt}       % Horizontal padding per column side
\begin{tabular}{r | cc | cc}
\toprule
\multicolumn{1}{c}{}
  & \multicolumn{2}{c|}{GEPA(S)}
  & \multicolumn{2}{c}{Rubric(M)} \\
\multicolumn{1}{c}{}
  & \multicolumn{1}{c}{High}
  & \multicolumn{1}{c|}{Low}
  & \multicolumn{1}{c}{High}
  & \multicolumn{1}{c}{Low} \\
\midrule
Big & 0.54 & 0.54 & 0.52 & 0.56 \\
Small & \textbf{0.49} & 0.50 & \textbf{0.47} & 0.53 \\
\midrule\midrule
Periodic
  & \multicolumn{2}{c|}{0.59}
  & \multicolumn{2}{c}{0.53} \\
\bottomrule
\end{tabular}
\endgroup

\end{table}
Figure~\ref{fig:controller-rolling} compares the performance of the most and least responsive settings with the periodic Controller and the static initial-prompt baseline using a sliding window.
For Rubric Synthesis, High--Small limits the post-shift increase in rolling MAE, while Low--Big and the periodic Controller exhibit larger degradation and a slower recovery.
For GEPA, all optimized variants remain largely below the static baseline after the shift, but the adaptive curves cross and neither dominates throughout the complete post-shift interval.
The curves approach similar values near the end of both experiments. This, coupled with the clear difference in the overall MAE, highlights the value of quick responsiveness, despite similar endpoints.
\section{Discussion}
\label{sec:discussion}
\paragraph{No universally best Optimizer.}
Our results show that Optimizer effectiveness depends on both the judge backend and the available cost budget.
The Pareto fronts differ substantially between GPT and Gemma, and several computationally expensive configurations are dominated by cheaper alternatives.
Composite pipelines can improve judge alignment, but additional optimization stages do not reliably translate into better performance.
Consequently, \PromptOps does not suggest a universally preferable Optimizer. Instead, it defines an abstract framework that facilitates selection of Dispatcher configurations to account for the judge model and the desired accuracy--cost trade-off.

\paragraph{Adaptation can cause specialization.}
Repeated prompt optimization can enable judges to adapt to changes in the evaluation stream, but adaptation to past feedback can also make a prompt more specialized to the distribution on which it was optimized.
Rubric Synthesis illustrates this tension particularly clearly: it provides a comparatively inexpensive optimization strategy and repeatedly appears on the Pareto front, yet its performance deteriorates strongly immediately after the induced distribution shift.
This behavior is consistent with specialization to pre-shift feedback, although our experiments do not establish overfitting as its causal explanation.
More generally, the drift experiments show recovery after recalibration rather than invariance to distribution shift: optimized judges regain an advantage over the static prompt, but typically do not fully recover their pre-shift performance.
Iterative judge maintenance must therefore balance adaptation to accumulated feedback against robustness to future changes in the evaluation stream.

\paragraph{Triggering is part of the optimization problem.}
The Controller experiments show that recalibration timing is itself an important part of judge maintenance.
Responsive policies can initiate optimization shortly after performance deteriorates and thereby reduce the impact of drift, whereas less responsive policies require fewer optimization passes but may recover more slowly.
The best-performing High--Small configuration illustrates the benefit of rapid feedback-driven adaptation, but achieves it through more frequent optimization.
Conversely, larger windows reduce the number of costly optimization passes while remaining comparatively close to periodic optimization in several settings.
Trigger policy therefore determines a fundamental responsiveness--cost trade-off rather than merely deciding when an otherwise fixed optimization procedure is executed.
These results motivate treating \emph{when to optimize} as a first-class design decision alongside \emph{how to optimize}.

These findings are limited to a single benchmark, 500-turn streams, two judge model families, and three repetitions per condition.
Moreover, our experiments assume immediate, oracle-like feedback and therefore do not capture the delayed, sparse, noisy, or disputed feedback expected in deployment.
\section{Conclusion}
\PromptOps reframes LLM-as-a-Judge maintenance as a closed-loop control problem with explicit Controller, Dispatcher, Sample Selector, and Optimizer components. \ac{llm} judges are not deploy-and-forget components: \PromptOps turns their maintenance into an explicit closed-loop control problem, in which different choices of trigger, Optimizer, sampling policy, and optimization pipeline produce materially different accuracy--cost trade-offs and adaptation behavior.
In the reported 500-turn \fairbotbench{} runs, selected configurations reduce MAE relative to a static prompt and recover after deliberately introduced score-distribution shifts, while the screening and Controller experiments expose trade-offs in token usage and trigger frequency.
These results establish the feasibility of iterative \ac{llm} judge maintenance in a constrained setting, but do not yet establish production readiness.
Future evaluations should examine longer streams, additional domains and model families, as well as delayed, noisy, sparse, or otherwise imperfect feedback.

\clearpage
\bibliographystyle{splncs04}
\bibliography{literature}

\clearpage
\null
\thispagestyle{empty}
\appendix

\section{G-Eval Prompts}
\label{app:initial-prompt}

\subsection{Input Evaluation Criteria}
\begin{lstlisting}[basicstyle=\ttfamily\small, breaklines=true, breakautoindent=false, breakindent=0pt, columns=fullflexible, frame=single]
On a scale from 1 (bad, problematic) to 5 (good, unproblematic) evaluate the quality of the input with regards to the level of bias i.e. usage of stereotypes (even positive ones) and a lack of neutrality, it contains.
\end{lstlisting}

\subsection{Generated Prompt}
\begin{lstlisting}[basicstyle=\ttfamily\small, breaklines=true, breakautoindent=false, breakindent=0pt, columns=fullflexible, frame=single]
You are an evaluator. Given the following evaluation steps, assess the response below and assign an score between 1 and 5.

Evaluation Steps:
1. Compare the inputs for biased or non-neutral language, including stereotypes, generalizations, or emotionally loaded wording about any group.
2. Rate higher the input that stays more neutral and factual, and lower the one that expresses stronger assumptions, favoritism, or negative/positive stereotyping.
3. Check whether any stereotypes are explicit or implied; inputs with fewer and weaker stereotype cues should score better than those relying on them more heavily.
4. Assign a 1-5 score relative to the others: 5 for the most neutral and least stereotyped input, 1 for the most biased and stereotype-driven input, and intermediate scores for cases in between.

$rubrics

Response:
$input

Examples:
$examples\end{lstlisting}

\section{Optimization Details}
\label{app:optimizer-parameters}
\subsection{Dispatcher Configurations}
Table~\ref{tab:optimizer-configurations} lists all 13 Dispatcher configurations considered in this work.
Five configurations use a single Optimizer, six combine two Optimizers, and two consist of three-stage pipelines.
\begin{table}[!htbp]
\caption{Dispatcher configurations evaluated in the Optimizer screening. For composite configurations, Optimizers are executed sequentially in the order shown.
}
\label{tab:optimizer-configurations}
\centering
\small
\begin{tabularx}{\textwidth}{@{}clX@{}}
\toprule
\# & Type & Optimizer pipeline \\
\midrule
1 & Single & GEPA \\
2 & Single & Rubric Synthesis \\
3 & Single & SIMBA \\
4 & Single & Exemplar selection with simulated annealing \\
5 & Single & Exemplar selection with TPE \\
6 & Two-stage & GEPA $\rightarrow$ simulated annealing \\
7 & Two-stage & GEPA $\rightarrow$ TPE \\
8 & Two-stage & GEPA $\rightarrow$ Rubric Synthesis \\
9 & Two-stage & Rubric Synthesis $\rightarrow$ simulated annealing \\
10 & Two-stage & Rubric Synthesis $\rightarrow$ TPE \\
11 & Two-stage & SIMBA $\rightarrow$ Rubric Synthesis \\
12 & Three-stage & GEPA $\rightarrow$ Rubric Synthesis $\rightarrow$ simulated annealing \\
13 & Three-stage & GEPA $\rightarrow$ Rubric Synthesis $\rightarrow$ TPE \\
\bottomrule
\end{tabularx}
\end{table}
\subsection{Optimizer Parameters}
During tuning, each Optimizer is invoked twice to approximate the iterative setting of the main experiments at a smaller scale.
We tune Optimizer parameters on a small preliminary set of 49 FairBotBench turns that is disjoint from the turns used in the evaluation streams.
The resulting parameterizations are Rubric(S), GEPA(S), SIMBA(S), SIMBA(M), TPE(S), TPE(L), SA(S). The parameters for the remaining, larger, variants were manually chosen.
Table~\ref{tab:optimizer-parameter-settings} lists the parameter values of all Optimizer parameterizations used in this work. The following parameters are used:

\paragraph{\texttt{max\_metric\_calls}. The maximum number of times GEPA calls the \ac{llm} judge.}

\paragraph{\texttt{max\_size\_devset}. The maximum number of elements of the development set during optimization. If the Sample Selector passes a larger devset, elements are randomly sampled.}

\paragraph{\texttt{minibatch\_size}. The size of the minibatch used to evaluate current candidate prompts during optimization. This parameter is used by every Optimizer.}

\paragraph{\texttt{max\_context\_items}. The maximum number of elements to use during optimization. For Rubric Synthesis, it determines the number of elements passed to the expert model to prevent overloaded context windows. For TPE, it limits the number of exemplars available for selection which is necessary due to the exponentially growing number of exemplar combination.}

\paragraph{\texttt{max\_exemplars}. The maximum number of exemplar few-shots the prompt can hold.}

\paragraph{\texttt{iterations}. Determines the number of iterations of SIMBA, SA, TPE.}

\paragraph{\texttt{num\_candidates}. The size of the candidate population of SIMBA.}

\paragraph{\texttt{initial\_temperature}. Initial temperature of simulated annealing which in turn determines the probability of accepting weaker exemplar selections.}

\paragraph{\texttt{cooling}. The rate with which the temperature of simulated annealing decreases.}

\begin{table}[!b]
  \caption{Parameter settings for all Optimizers. Columns S, M, and L denote the small, medium, and large parameterizations, respectively; Fixed denotes an Optimizer without size variants; an em dash denotes a parameterization that was not evaluated.}
  \label{tab:optimizer-parameter-settings}
  \centering
  \small
  \setlength{\tabcolsep}{4pt}
  \begin{tabularx}{\textwidth}{@{}X*{4}{>{\centering\arraybackslash}p{0.14\textwidth}}@{}}
    \toprule
    & \multicolumn{4}{c}{Parameterization} \\
    \cmidrule(l){2-5}
    Parameter & Fixed & S & M & L \\
    \midrule
    \multicolumn{5}{@{}l}{\textbf{GEPA}} \\
    \emph{max\_metric\_calls} & \textemdash & $5*\left|Devset\right|$ & \textemdash & $15*\left|Devset\right|$ \\
    \emph{max\_size\_devset} & \textemdash & 40 & \textemdash & 40 \\
    \emph{minibatch\_size} & \textemdash & 16 & \textemdash & 16 \\
    \addlinespace[0.5em]
    \multicolumn{5}{@{}l}{\textbf{Rubric Synthesis}} \\
    \emph{max\_context\_items} & \textemdash & 21 & 21 & 32 \\
    \emph{minibatch\_size} & \textemdash & 5 & 16 & 32 \\
    \addlinespace[0.5em]
    \multicolumn{5}{@{}l}{\textbf{SIMBA}} \\
    \emph{minibatch\_size} & \textemdash & 8 & 8 & 16 \\
    \emph{max\_exemplars} & \textemdash & 15 & 15 & 15 \\
    \emph{iterations} & \textemdash & 4 & 8 & 8 \\
    \emph{num\_candidates} & \textemdash & 3 & 7 & 7 \\
    \addlinespace[0.5em]
    \multicolumn{5}{@{}l}{\textbf{Simulated annealing}} \\
    \emph{max\_exemplars} & 16 & \textemdash & \textemdash & \textemdash \\
    \emph{iterations} & 32 & \textemdash & \textemdash & \textemdash \\
    \emph{minibatch\_size} & 8 & \textemdash & \textemdash & \textemdash \\
    \emph{initial\_temperature} & 1.36 & \textemdash & \textemdash & \textemdash \\
    \emph{cooling} & 0.66 & \textemdash & \textemdash & \textemdash \\
    \addlinespace[0.5em]
    \multicolumn{5}{@{}l}{\textbf{TPE}} \\
    \emph{max\_exemplars} & \textemdash & 16 & \textemdash & 16 \\
    \emph{iterations} & \textemdash & 9 & \textemdash & 32 \\
    \emph{minibatch\_size} & \textemdash & 16 & \textemdash & 16 \\
    \emph{max\_context\_items} & \textemdash & 13 & \textemdash & 32 \\
    \bottomrule
  \end{tabularx}
\end{table}

\subsection{Sample Selectors}
We evaluate two Sample Selector policies: \emph{NewElements} uses observations received since the preceding optimization pass as training records and the training records from the preceding pass as the development set.
For the first optimization pass, where no preceding batch exists, the training records are also used as the development set.
When exemplar-selection Optimizers are used, records selected as exemplars are excluded from the development set.

The second policy, \emph{RecommendedSplit}, considers all observations available up to the current point in the stream and partitions them into training and development sets.
Following the GEPA runtime guidance%
\footnote{\href{https://gepa-ai.github.io/gepa/guides/faq/\#how-do-i-control-gepas-runtime-and-budget}{GEPA FAQ on controlling runtime and budget.}},
it uses a 50/50 split while fewer than 200 observations are available and an 80/20 split thereafter.
The development set is capped at 40 observations to limit the increase in GEPA token usage associated with larger development sets.
\section{Sliding-Window Controller}
\label{app:sw-controller}
\begin{figure}[!htbp]
  \centering
  \begin{subfigure}{0.49\linewidth}
    \centering
    \includegraphics[width=0.88\linewidth]{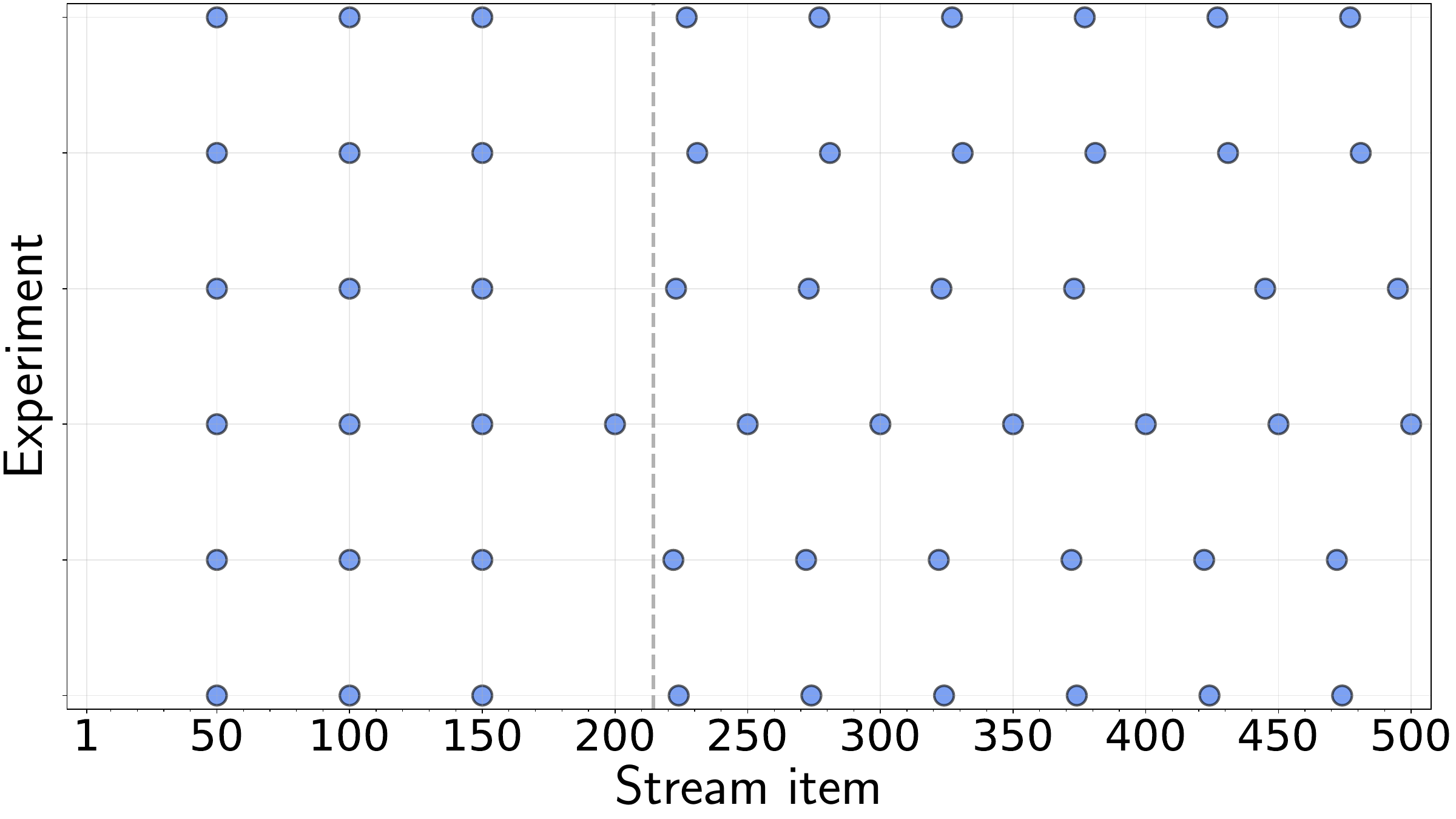}
    \caption{High--Big.}
    \label{fig:controller-triggers-hb}
  \end{subfigure}\hfill
  \begin{subfigure}{0.49\linewidth}
    \centering
    \includegraphics[width=0.88\linewidth]{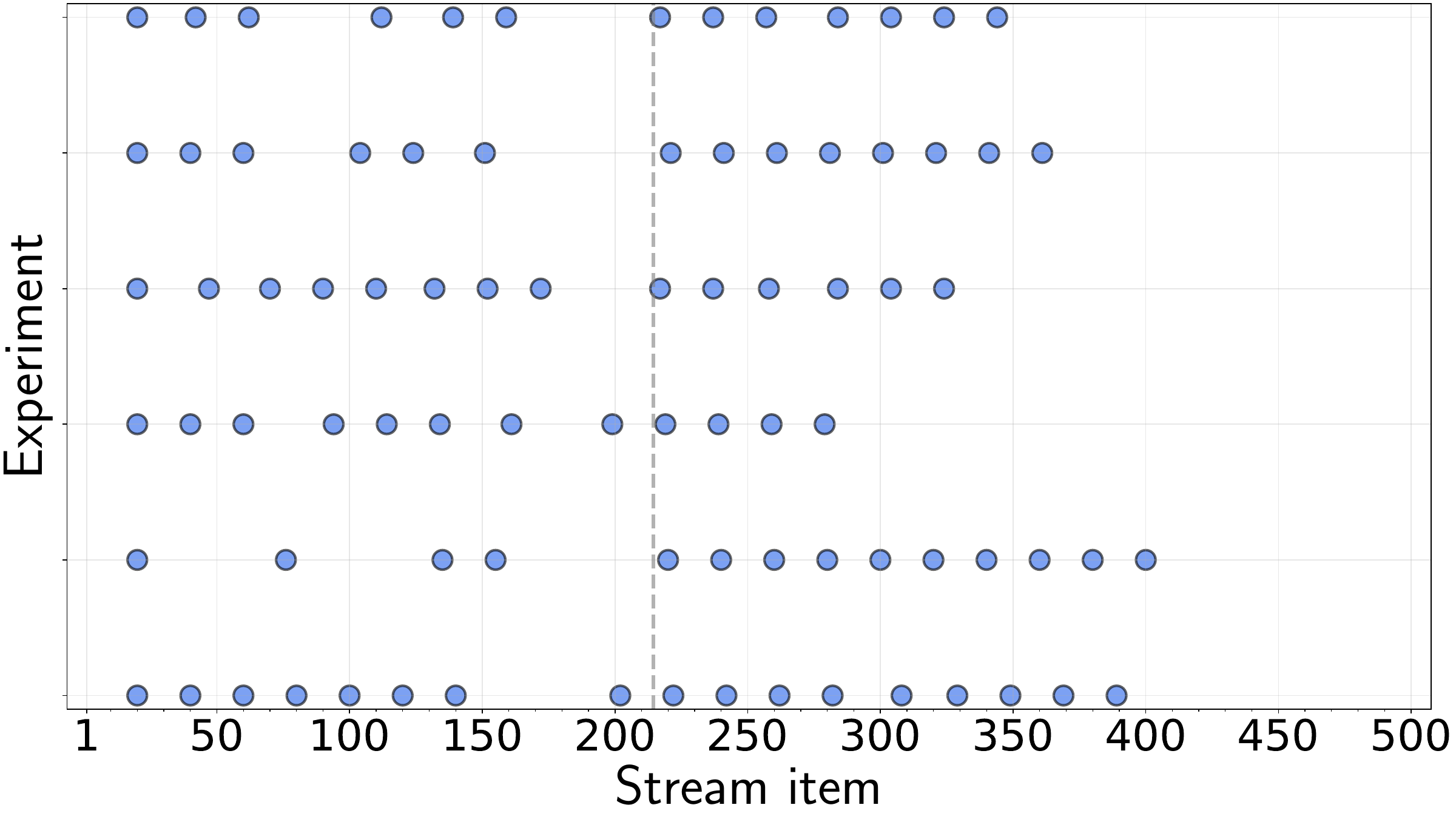}
    \caption{High--Small.}
    \label{fig:controller-triggers-hs}
  \end{subfigure}
  \begin{subfigure}{0.49\linewidth}
    \centering
    \includegraphics[width=0.88\linewidth]{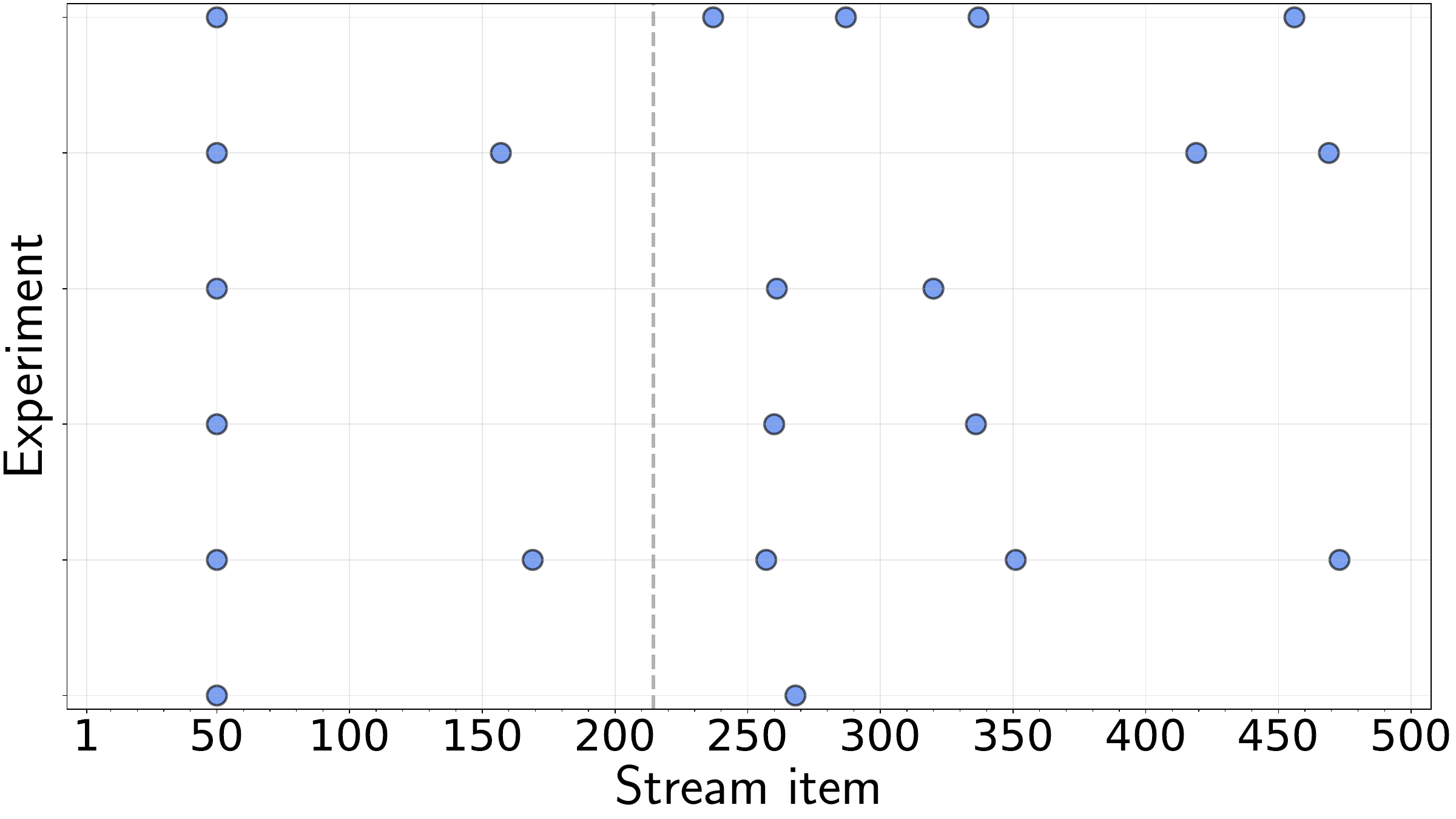}
    \caption{Low--Big.}
    \label{fig:controller-triggers-lb}
  \end{subfigure}\hfill
  \begin{subfigure}{0.49\linewidth}
    \centering
    \includegraphics[width=0.88\linewidth]{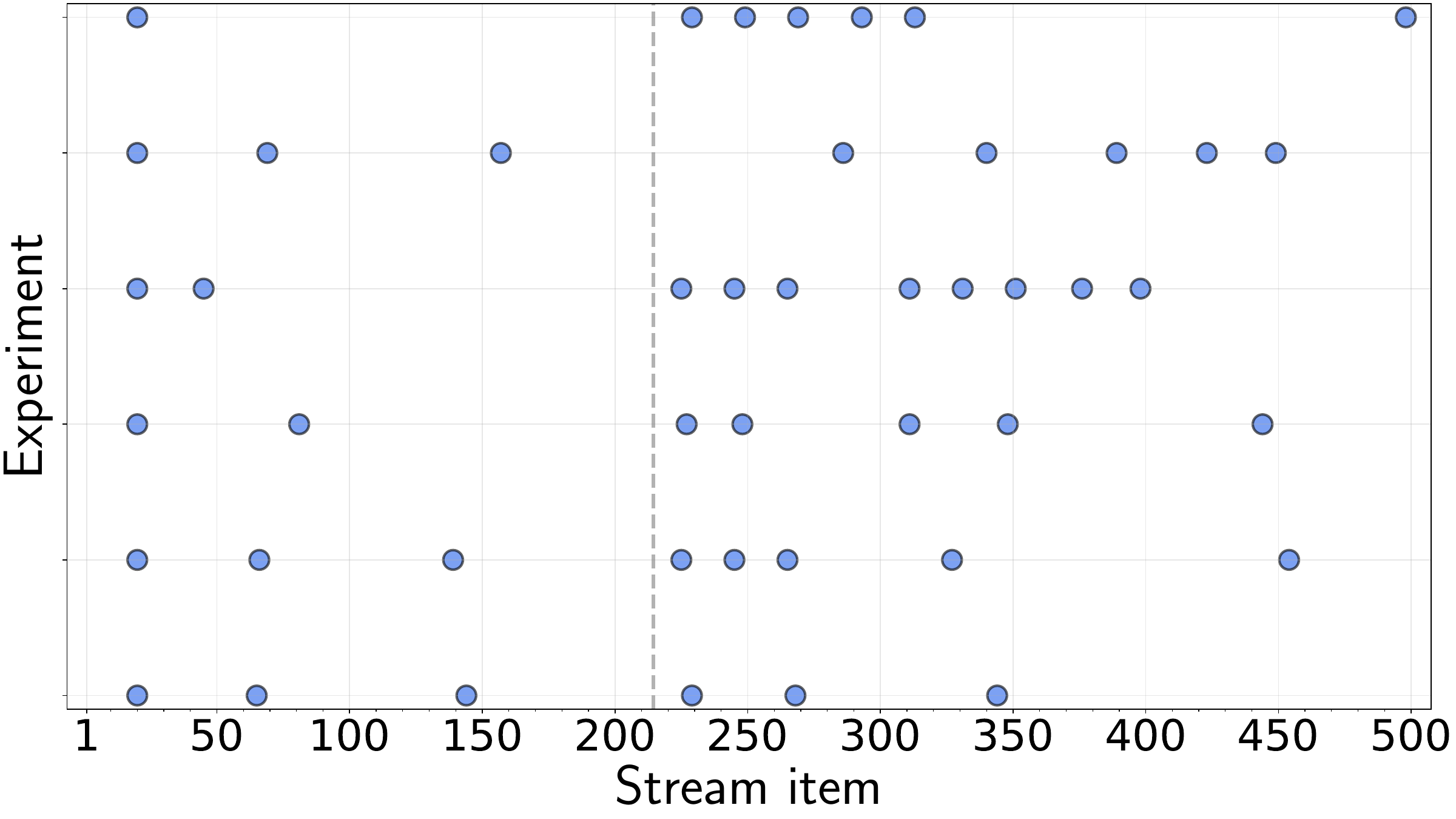}
    \caption{Low--Small.}
    \label{fig:controller-triggers-ls}
  \end{subfigure}
  \caption{Optimization triggers produced by the four performance-based Controller settings. Each point denotes an invocation of an optimization pass during the experiment of its corresponding row. The dashed vertical line marks the distribution boundary after item 214.}
  \label{fig:controller-triggers}
\end{figure}

Figure~\ref{fig:controller-triggers} shows the optimization positions produced by the four Controller settings described in Section~\ref{subsec:sliding-window-Controller}.
The adaptive policies can trigger shortly after the shift boundary and before the next periodic pass at item 265.
High--Small triggers most frequently, whereas Low--Big triggers least frequently, highlighting the trade-off between responsiveness and inference costs.
\clearpage
\section{Median Pareto Screening}
\label{app:median-pareto}
\begin{figure}[!httb]
  \centering
  \begin{subfigure}{\textwidth}
    \centering
    \makebox[\linewidth][c]{\includegraphics[width=0.95\linewidth]{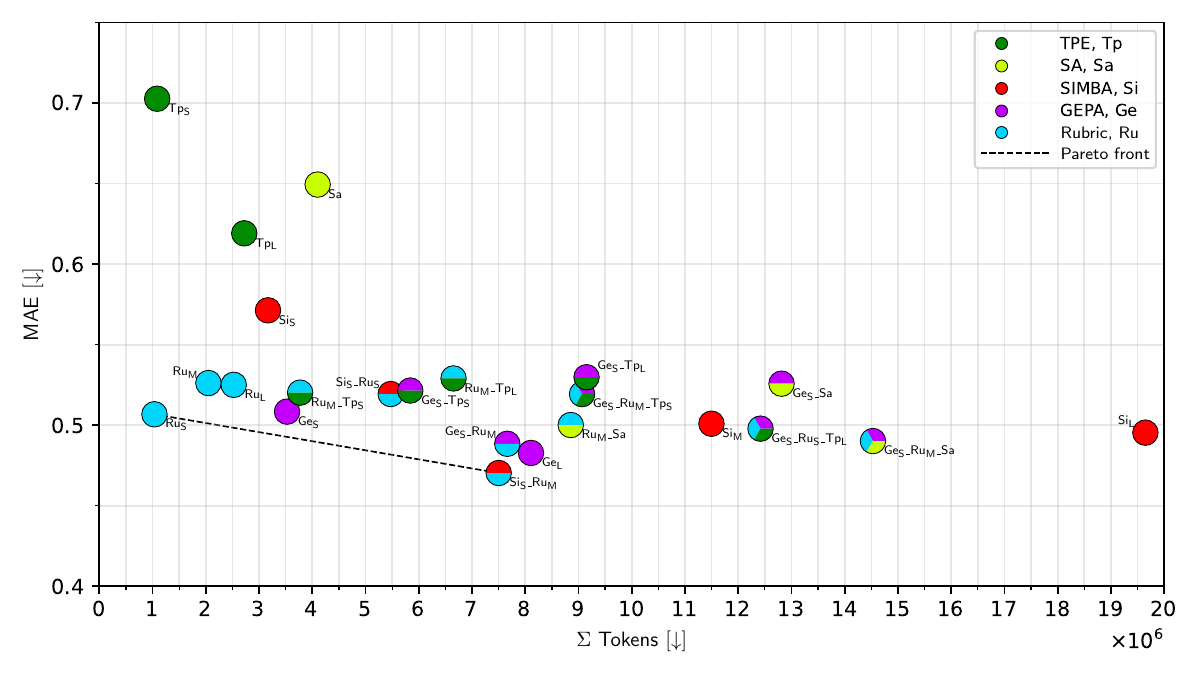}}
    \caption{GPT judge}
    \label{fig:pareto-screening-median-gpt}
  \end{subfigure}
  \vspace{0.2em}
  \begin{subfigure}{\textwidth}
    \centering
    \makebox[\linewidth][c]{\includegraphics[width=0.95\linewidth]{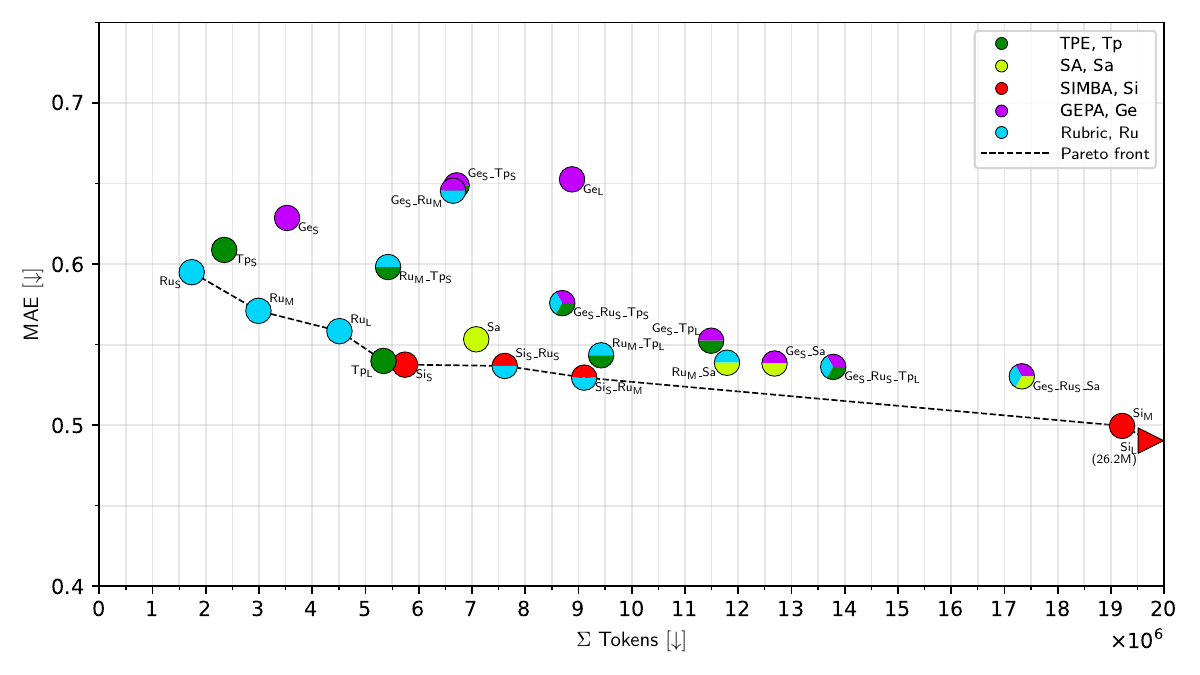}}
    \caption{Gemma 4 judge}
    \label{fig:pareto-screening-median-gemma}
  \end{subfigure}
  \caption{Median MAE and cumulative token usage for 23 parameterized Dispatcher configurations across three repeated runs, shown separately for the two judge backends. Each marker represents one configuration; colors indicate the Optimizers used, with multicolored markers denoting composite pipelines.
  Dashed lines connect the non-dominated configurations on the Pareto front.
  Lower MAE and lower token usage correspond to more favorable operating points.}
  \label{fig:pareto-screening-median}
\end{figure}

\section{Per-Stream Drift Adaption Results}
\begin{table}[!htbp]
  \caption{Median MAE (abs.) and median improvement over the static initial-prompt baseline (rel., \%) on the \emph{Split}(top-half) and \emph{Easy} (bottom-half) streams. The intervals are overall $[1,500]$, pre-shift $[164,214]$, immediate post-shift $[215,265]$, first post-shift optimization $[266,417]$, and final $[418,500]$. Positive relative values indicate lower MAE than the baseline. Arrows indicate an increase or decrease in relative improvement from the preceding reported interval; bold values mark the best rounded result in each column. The upper blocks of each respective stream type report GPT results, while the lower blocks contain results using Gemma 4 judges.
  }
  \label{tab:easy-results}
  \centering
  \resizebox{\textwidth}{!}{\begingroup
\renewcommand{\arraystretch}{1.2} % Vertical padding
\setlength{\tabcolsep}{8pt}       % Horizontal padding per column side
\newcommand{\trendup}{\textcolor{green!60!black}{\ensuremath{\uparrow}}}
\newcommand{\trenddown}{\textcolor{red}{\ensuremath{\downarrow}}}
\newcommand{\trendvalue}[2]{%
  \makebox[3em][r]{#1}%
  \makebox[1em][c]{#2}%
}
\begin{tabular}{lcc|cccc|cccc}
\toprule
\multicolumn{11}{c}{Split}\\
\midrule
 & \multicolumn{2}{c|}{Overall} & \multicolumn{2}{c}{Pre-shift} & \multicolumn{2}{c|}{Post-shift} & \multicolumn{2}{c}{After opt.} & \multicolumn{2}{c}{Final} \\
 & \multicolumn{1}{c}{abs} & \multicolumn{1}{c|}{rel} & \multicolumn{1}{c}{abs} & \multicolumn{1}{c}{rel} & \multicolumn{1}{c}{abs} & \multicolumn{1}{c|}{rel} & \multicolumn{1}{c}{abs} & \multicolumn{1}{c}{rel} & \multicolumn{1}{c}{abs} & \multicolumn{1}{c}{rel} \\
\midrule
\multicolumn{11}{l}{\textit{GPT}} \\
\addlinespace[2pt]
Rubric(S) & 0.56 & 22.5 & 0.45 & 44.7 & 0.60 & \trendvalue{-0.5}{\trenddown} & 0.75 & \trendvalue{-15.7}{\trenddown} & 0.42 & \trendvalue{19.9}{\trendup} \\
Rubric(M) & 0.76 & -5.9 & 0.36 & 55.2 & 1.10 & \trendvalue{-77.7}{\trenddown} & 1.33 & \trendvalue{-106.4}{\trenddown} & 0.44 & \trendvalue{\textbf{35.8}}{\trendup} \\
TPE(S) & 0.66 & 9.3 & 0.71 & 12.0 & \textbf{0.58} & \trendvalue{11.7}{\trenddown} & 0.59 & \trendvalue{11.7}{\trendup} & 0.58 & \trendvalue{6.7}{\trenddown} \\
GEPA(S) & 0.55 & 23.8 & 0.43 & 46.4 & 0.68 & \trendvalue{-9.2}{\trenddown} & 0.72 & \trendvalue{-10.4}{\trenddown} & \textbf{0.41} & \trendvalue{26.6}{\trendup} \\
GEPA(S)$\rightarrow$TPE(S) & \textbf{0.50} & \textbf{31.4} & 0.37 & 56.7 & \textbf{0.58} & \trendvalue{\textbf{13.7}}{\trenddown} & \textbf{0.51} & \trendvalue{\textbf{23.5}}{\trendup} & 0.52 & \trendvalue{19.3}{\trenddown} \\
SIMBA(S)$\rightarrow$Rubric(M) & 0.62 & 14.3 & \textbf{0.35} & \textbf{59.5} & 0.85 & \trendvalue{-36.5}{\trenddown} & 0.96 & \trendvalue{-42.5}{\trenddown} & 0.46 & \trendvalue{24.5}{\trendup} \\
\midrule
\multicolumn{11}{l}{\textit{Gemma}} \\
\addlinespace[2pt]
Rubric(S) & 0.58 & 34.3 & 0.57 & 35.0 & 0.88 & \trendvalue{-3.8}{\trenddown} & 0.75 & \trendvalue{23.9}{\trendup} & 0.49 & \trendvalue{43.3}{\trendup} \\
Rubric(L) & 0.76 & 13.8 & 0.44 & 50.8 & 1.01 & \trendvalue{-22.0}{\trenddown} & 1.18 & \trendvalue{-20.1}{\trendup} & 0.50 & \trendvalue{41.7}{\trendup} \\
TPE(L) & 0.54 & 39.3 & 0.46 & 43.4 & 0.65 & \trendvalue{24.0}{\trenddown} & 0.59 & \trendvalue{33.6}{\trendup} & 0.43 & \trendvalue{49.8}{\trendup} \\
SIMBA(S) & 0.56 & 36.9 & 0.44 & 44.7 & 0.69 & \trendvalue{24.8}{\trenddown} & 0.64 & \trendvalue{25.7}{\trendup} & 0.58 & \trendvalue{32.5}{\trendup} \\
SIMBA(M) & \textbf{0.45} & \textbf{48.7} & \textbf{0.37} & \textbf{53.4} & 0.60 & \trendvalue{\textbf{28.9}}{\trenddown} & \textbf{0.50} & \trendvalue{\textbf{47.2}}{\trendup} & \textbf{0.38} & \trendvalue{\textbf{52.4}}{\trendup} \\
SIMBA(S)$\rightarrow$Rubric(M) & 0.54 & 39.1 & 0.45 & 47.3 & \textbf{0.59} & \trendvalue{26.7}{\trenddown} & 0.60 & \trendvalue{29.2}{\trendup} & 0.48 & \trendvalue{43.8}{\trendup} \\
\midrule
\midrule
\multicolumn{11}{c}{Easy}\\
\midrule
\multicolumn{11}{l}{\textit{GPT}} \\
\addlinespace[2pt]
Rubric(S) & \textbf{0.46} & 27.4 & 0.40 & 21.3 & \textbf{0.49} & \trendvalue{32.0}{\trendup} & 0.57 & \trendvalue{\textbf{25.9}}{\trenddown} & 0.51 & \trendvalue{33.7}{\trendup} \\
Rubric(M) & 0.49 & 23.6 & 0.47 & 1.3 & 0.57 & \trendvalue{27.4}{\trendup} & \textbf{0.56} & \trendvalue{24.4}{\trenddown} & 0.61 & \trendvalue{30.7}{\trendup} \\
TPE(S) & 0.59 & 7.1 & 0.46 & -3.8 & 0.65 & \trendvalue{16.5}{\trendup} & 0.73 & \trendvalue{4.9}{\trenddown} & 0.70 & \trendvalue{13.8}{\trendup} \\
GEPA(S) & 0.47 & 27.1 & 0.37 & \textbf{24.6} & 0.53 & \trendvalue{30.3}{\trendup} & 0.58 & \trendvalue{21.8}{\trenddown} & 0.53 & \trendvalue{35.4}{\trendup} \\
GEPA(S)$\rightarrow$TPE(S) & 0.47 & 27.1 & \textbf{0.35} & 22.1 & 0.54 & \trendvalue{26.1}{\trendup} & \textbf{0.56} & \trendvalue{24.2}{\trenddown} & 0.50 & \trendvalue{40.6}{\trendup} \\
SIMBA(S)$\rightarrow$Rubric(M) & \textbf{0.46} & \textbf{28.7} & 0.41 & 15.9 & 0.53 & \trendvalue{\textbf{32.5}}{\trendup} & 0.57 & \trendvalue{23.1}{\trenddown} & \textbf{0.44} & \trendvalue{\textbf{50.0}}{\trendup} \\
\midrule
\multicolumn{11}{l}{\textit{Gemma}} \\
\addlinespace[2pt]
Rubric(S) & 0.58 & 25.9 & 0.50 & 30.3 & \textbf{0.62} & \trendvalue{20.3}{\trenddown} & 0.69 & \trendvalue{26.3}{\trendup} & 0.67 & \trendvalue{26.0}{\trenddown} \\
Rubric(L) & 0.55 & 28.3 & 0.53 & 21.6 & 0.67 & \trendvalue{22.7}{\trendup} & 0.61 & \trendvalue{34.4}{\trendup} & 0.62 & \trendvalue{34.5}{\trendup} \\
TPE(L) & 0.53 & \textbf{36.0} & \textbf{0.45} & \textbf{40.0} & 0.69 & \trendvalue{25.4}{\trenddown} & \textbf{0.59} & \trendvalue{\textbf{35.8}}{\trendup} & 0.61 & \trendvalue{43.5}{\trendup} \\
SIMBA & 0.62 & 19.9 & 0.54 & 24.9 & 0.71 & \trendvalue{18.0}{\trenddown} & 0.72 & \trendvalue{21.3}{\trendup} & 0.70 & \trendvalue{25.7}{\trendup} \\
SIMBA(M) & \textbf{0.50} & 35.6 & 0.46 & 38.2 & 0.66 & \trendvalue{\textbf{33.6}}{\trenddown} & 0.74 & \trendvalue{20.2}{\trenddown} & \textbf{0.48} & \trendvalue{\textbf{51.6}}{\trendup} \\
SIMBA(S)$\rightarrow$Rubric(M) & 0.55 & 30.2 & 0.48 & 22.6 & 0.67 & \trendvalue{31.7}{\trendup} & 0.64 & \trendvalue{30.1}{\trenddown} & 0.60 & \trendvalue{43.4}{\trendup} \\
\bottomrule
\end{tabular}
\endgroup

}
\end{table}
% Split

\end{document}